\documentclass[10pt,twocolumn,letterpaper]{article}

\usepackage{iccv}
\usepackage{times}
\usepackage{epsfig}
\usepackage{graphicx}
\usepackage{amsmath}
\usepackage{amssymb}
\usepackage{booktabs}
\usepackage{multirow}
\usepackage{dsfont}
\usepackage{subfig}

\usepackage[pagebackref=true,breaklinks=true,colorlinks,bookmarks=false]{hyperref}

\iccvfinalcopy 

\def\iccvPaperID{10913} 

\newcommand{\specialcell}[2][c]{%
  \begin{tabular}[#1]{@{}c@{}}#2\end{tabular}}
\DeclareMathOperator*{\argmax}{arg\,max}
\ificcvfinal\fi

\begin{document}

\title{Domain generalization of 3D semantic segmentation in autonomous driving}

\author{Jules Sanchez$^1$
\and 
Jean-Emmanuel Deschaud$^1$
\and 
François Goulette$^{1,2}$
\and 
$^1$\textit{Centre for Robotics, Mines Paris - PSL,}\\ \textit{PSL University,} \\
75006 Paris, France \\
\small{firstname.surname@minesparis.psl.eu}
\and
$^2$\textit{U2IS, ENSTA Paris,}\\ \textit{Institut Polytechnique de Paris,} \\
91120 Palaiseau, France \\
\small{firstname.surname@ensta-paris.fr}}

\maketitle

\begin{abstract}

Using deep learning, 3D autonomous driving semantic segmentation has become a well-studied subject, with methods that can reach very high performance. Nonetheless, because of the limited size of the training datasets, these models cannot see every type of object and scene found in real-world applications. The ability to be reliable in these various unknown environments is called \textup{domain generalization}.

Despite its importance, domain generalization is relatively unexplored in the case of 3D autonomous driving semantic segmentation. To fill this gap, this paper presents the first benchmark for this application by testing state-of-the-art methods and discussing the difficulty of tackling Laser Imaging Detection and Ranging (LiDAR) domain shifts.

We also propose the first method designed to address this domain generalization, which we call 3DLabelProp. This method relies on leveraging the geometry and sequentiality of the LiDAR data to enhance its generalization performances by working on partially accumulated point clouds. It reaches a mean Intersection over Union (mIoU) of 50.4\% on SemanticPOSS and of 55.2\% on PandaSet solid-state LiDAR while being trained only on SemanticKITTI, making it the state-of-the-art method for generalization (+5\% and +33\% better, respectively, than the second best method).

The code for this method is available on GitHub: \small\url{https://github.com/JulesSanchez/3DLabelProp}.

\end{abstract}

\section{Introduction}
\label{sec:intro}
Since the release of SemanticKITTI \cite{behley2019iccv}, 3D semantic segmentation for autonomous driving has been a field of growing interest, resulting in the emergence of a large variety of open-source datasets \cite{nuscenes2019,poss,360} and high-performing methods \cite{Zhou2020Cylinder3DAE,spvnas}.

Using LiDAR data instead of the typical 2D images obtained from cameras has several advantages. First, LiDAR data are resilient to some sources of 2D domain shifts, such as the change in illumination. Furthermore, these data yield precise geometric information and reliable distance information relative to the objects in scenes (down to a few centimeters in error).
\begin{figure}[t]
    \centering
    \includegraphics[width=\linewidth]{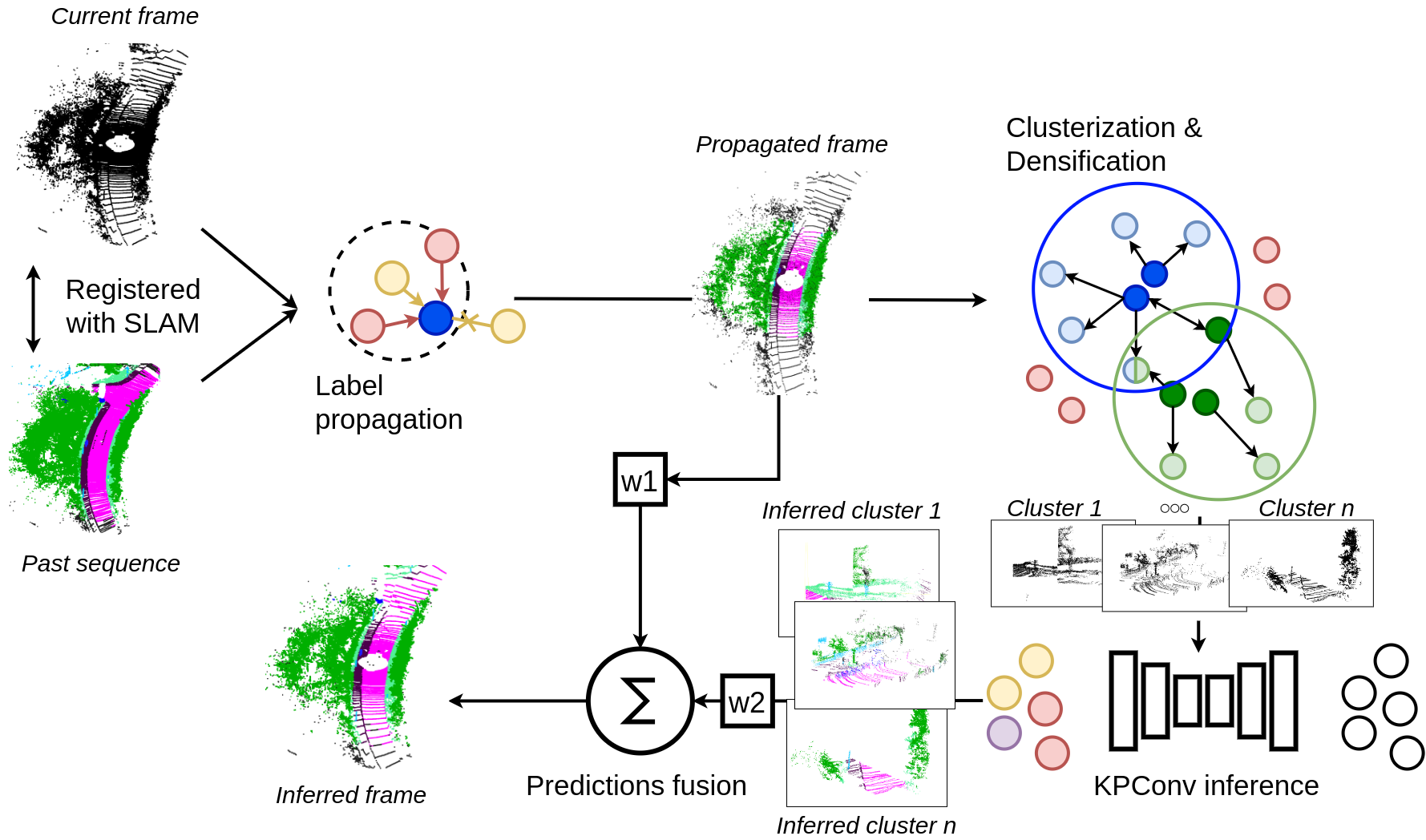}
    \caption{Illustration of the 3DLabelProp algorithm. With the past inferred sequence and the current registered LiDAR frame given as inputs, 3DLabelProp geometrically propagates the static object labels. Then, it clusterizes the current scan and populates it with points from the past sequence to feed small but dense point clouds to the KPConv module. Finally, predictions from the geometric propagation and deep learning module are combined to obtain the inferred LiDAR frame.}
    \label{fig:fastsegnet}
\end{figure}
Nevertheless, deep learning methods dealing with LiDAR data are still sensitive to domain shifts such as scene-type change (urban vs. suburban) or sensor change \cite{9341508}. This observation is especially relevant for autonomous driving semantic segmentation because the source of the domain shift varies and originates from hardware, such as the number of fibers in the LiDAR and its positioning on the vehicles; from firmware such as reflectivity calculations; and from the scene itself. LiDAR data are also costly to acquire and label (1,700 hours of labelization for SemanticKITTI \cite{behley2019iccv}). The accumulation of these issues has clarified the need for domain adaptation and domain generalization methods to, on the one hand leverage the availability of synthetic data \cite{deschaud2021kitticarla,xiao2022transfer} and, on the other hand, perform well on unseen data because it is not feasible to cover every scene type and corner case.

Data adaptation aims to bridge the domain shift with a learning transferable model; hence, data adaptation-methods usually train a model on source data and then perform an adaptation step, for which they have access to either sparsely labeled or fully unlabeled target data. Conversely, domain generalization methods work in a setup where the target data are fully unavailable.

Although domain adaptation has gained a lot of traction in 3D scene understanding, domain generalization is still relatively unexplored. Furthermore, the disparity between the label sets of various available datasets makes it difficult to compare methods across datasets. In this work, we propose a simple method to evaluate domain generalization performance. With it, we benchmark current state-of-the-art approaches for autonomous driving semantic segmentation. 

Furthermore, we propose a novel method to perform semantic segmentation while improving generalization performances. Specifically, we develop a domain alignment method, which accumulates the sequences of LiDAR frames rather than examining scans individually, as is usually done. Although working on such point clouds is slow, we propose several geometry-based mechanisms to speed up the method by propagating the labels of static objects to new points and processing small clusters of points rather than the full point cloud. This method is called \textit{3DLabelProp}.

We can summarize our contributions as follows:
\begin{itemize}
    \item We properly introduce the domain generalization problem of 3D semantic segmentation in the autonomous driving field and evaluate current state-of-the-art methods.
    \item We propose a domain generalization-oriented method for autonomous driving semantic segmentation called 3DLabelProp (\autoref{fig:fastsegnet}).

\end{itemize}

\section{Related works}
\label{sec:sota}
\subsection{Autonomous driving 3D semantic segmentation}

To tackle the task of 3D autonomous driving semantic segmentation and its speed requirements, older permutations invariant methods were omitted \cite{Tat2018,10.1007/978-3-030-01237-3_6} in favor of methods reordering point clouds into structured representations. These methods are divided in two main approaches: projecting the point clouds in 2D with a range-based image \cite{8967762}\cite{squeezesegv3} and bird's-eye-view image \cite{Zhang_2020_CVPR} or leveraging sparse convolution-based architectures as introduced by MinkowskiNet \cite{mink}, under the name Sparse Residual U-Net (SRUNet). SRUNets have been used as the backbone for more refined methods \cite{mink,spvnas,Zhou2020Cylinder3DAE,Xu2021RPVNetAD}, which are the best-performing methods on the various benchmarks. Some multimodal methods linking images and LiDAR are also starting to appear \cite{yan20222dpass} but are outside the scope of the present article.

Because of the computation time constraint, most of these methods process frames one by one, discarding the sequentiality of the autonomous driving data. Recent works have started to exploit this for moving object segmentation \cite{9796597} or for full segmentation thanks to flow estimation \cite{9010250}, 4D representations \cite{loiseau22online,fan2021pstnet}, or 4D partial aggregation \cite{aygun20214d,kreuzberg2022stop}. 

Aggregation-based methods are similar to offline outdoor semantic segmentation, which is a well-studied field reaching great performances \cite{thomas2019KPConv}. Nevertheless, these methods are designed for offline processing and are inherently slow and unsuited for autonomous driving segmentation as such.

3DLabelProp is inspired by aggregation-based methods, with a focus on improving speed and generalization performances.

\subsection{Domain generalization for autonomous driving semantic segmentation}

Domain generalization in 2D has been thoroughly explored, working on training techniques such as meta-learning \cite{Li2018MLDG}, episodic training \cite{9008109}, and multitask learning \cite{8953760}; data augmentation \cite{NEURIPS2018_1d94108e,randconv2021}; or architecture design \cite{pan2018IBN-Net,8237429,9513542}. 

Meta-learning methods provide a different training algorithm rather than the typical approach, specifically in \cite{Li2018MLDG}, which simulates training and testing domains at each batch and ensures that the training and testing gradients are aligned. Episodic training \cite{9008109} pretrains several domain-specific feature extractors and then combines them at training time to make them robust against domain shifts. These methods rely on the availability of various source domains or the ability to simulate them at training time. Multitask learning aims to associate self-supervised or unsupervised tasks alongside the main one to learn better representations to be less sensitive to domain shifts. Data augmentation techniques provide a set of rules and methods to enhance the data at training time to diversify the available data \cite{9710159}, usually by modifying appearance and sensor information such as texture, brightness, and contrast \cite{randconv2021}; these techniques can also specialize in finding relevant and difficult samples of the available data \cite{NEURIPS2018_1d94108e}. Finally, architecture design methods proposed new neural blocks to improve domain generalization, such as normalization blocks, to align samples from different domains \cite{pan2018IBN-Net}.

A more thorough overview and discussions of the various domain generalization paradigms and algorithms can be found in \cite{9847099} \cite{9782500}.

In point cloud processing, domain generalization has been mainly evaluated and studied for 3D registration \cite{9896915} \cite{horache2021mssvconv} and object classification \cite{10.1007/978-3-031-19809-0_16,9577308}. For scene understanding, fewer studies have been released and have focused on object detection, such as 3D-VField \cite{9879166}, which proposes new data augmentation techniques by applying point cloud deformation. Besides these works, common LiDAR scene understanding works do not measure the domain generalization abilities of the various architectures.

While domain adaptation is different from domain generalization, unsupervised domain adaptation (UDA) is a closely related research area and is relevant to include here. We can distinguish sim-to-real UDA, which pretrains a model on synthetic data and adapts it on real data. UDA methods try to align the style of synthetic to real point clouds such as in SqueezeSeg \cite{wu2017squeezeseg} and ePointDA \cite{Zhao2021ePointDAAE}, even training GANs to learn and transfer scarcity and scene type \cite{xiao2022transfer}; there is also real-to-real domain adaptation, which employs alignment techniques such as learning completion on point clouds to train on this intermediate representations \cite{9578920} or resampling point clouds to align the target data with the source data \cite{9341508}.

Similar to \cite{9578920}, 3DLabelProp proposes finding a common representation of autonomous driving LiDAR data, namely the partially accumulated point clouds, such that the features of the target point clouds are similar to the features of the source point clouds.

\section{Domain generalization}

\subsection{Definition}

Domain generalization addresses the transferability of a model or an algorithm from one source to a target domain, without the availability of the target domain prior to the inference. Nonetheless, this unavailability of the target domain is not synonymous with the absence of priors on the target data. 

For autonomous driving LiDAR semantic segmentation, it can be assumed that point clouds will represent similar scenes, namely driving environments, hence representing similar objects, such as cars, pedestrians, and roads.

Consequently, we can identify three main domain shift components: 
\begin{itemize}
    \item Sensor shift, which stems from the resolution, setup, technology and model of the LiDAR sensor on the acquisition platform.
    \item Appearance shift, which originated from the different appearances of similar objects in different scenes, such as car sizes or vegetation variety. It also encapsulates weather and season change that can affect appearance of the objects such as tree losing their leaves in the winter, or fog adding noise to the acquisition.
    \item Scene shift, which is caused by different scenes displaying different objects in their environments; for example traffic lights are typical of an urban scene but not highway scenes. 
\end{itemize}

\begin{figure}[ht]
    \centering
    \includegraphics[width=\linewidth]{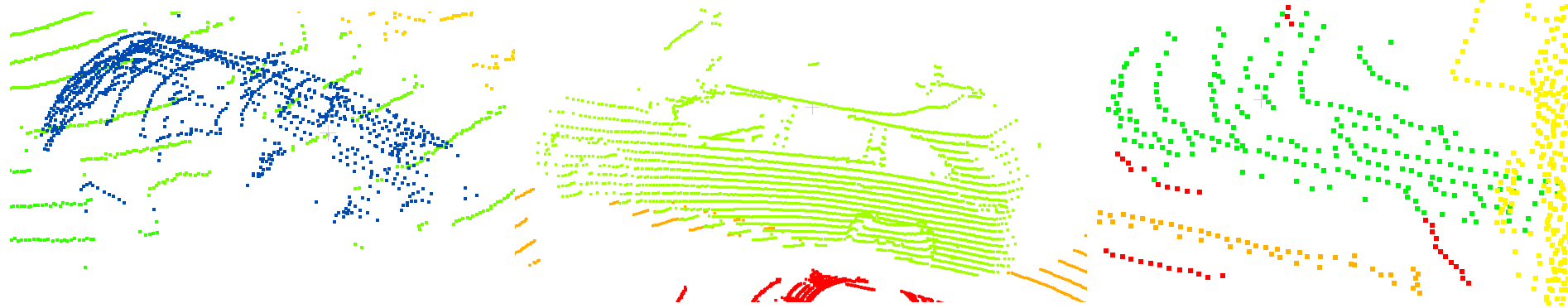}
    \caption{Example of different cars. From left to right: in SemanticKITTI, in nuScenes, and in SemanticPOSS.}
    \label{fig:shift_illustration}
\end{figure}

Examples of appearance and sensor shifts can be seen in \autoref{fig:shift_illustration}.

\subsection{Evaluation}
\label{sec:evaluation}

\subsubsection{Datasets}

\begin{table}[h]
    \scriptsize
    \centering
    \begin{tabular}{c|c|c}
        name & LiDAR technology & \# fibers \\ \hline
        SemanticKITTI & rotative  & 64 \\ \hline
        SemanticKITTI32 & rotative  & 32 \\ \hline
        Panda64 & rotative  & 64 \\ \hline
        PandaFF & solid-state  & 150 \\ \hline
        SemanticPOSS & rotative  & 40 \\ \hline
        nuScenes & rotative  & 32 \\ 
    \end{tabular}
    \caption{Sensor information for the used datasets.}
    \label{tab:sensor_info}
\end{table}
In order to benchmark the generalization performances of the state-of-the-art LiDAR semantic segmentation algorithms, we leverage several open-source datasets: SemanticKITTI \cite{behley2019iccv}, SemanticPOSS \cite{poss}, nuScenes \cite{nuscenes2019}, and PandaSet \cite{panda}. 

SemanticKITTI is a standard, widely used dataset in 3D Vision and is based on a 360-degree rotating LiDAR with 64 fibers. We will use it as our main training source dataset. To study sensor shift in a configuration without any other shifts, we introduce SemanticKITTI32, which is generated by removing one out of two fibers from the original SemanticKITTI dataset. 

We divide PandaSet into two sub-datasets, Panda64 (P64), which are the frames acquired by the 64-fiber LiDAR inside PandaSet, and PandaFF (PFF), which are the frames acquired by the front-facing LiDAR inside PandaSet. It is important to note that PandaFF is a dataset acquired by a solid-state LiDAR. 

SemanticPOSS is a dataset acquired at a Chinese university campus by a 360-degree rotating LiDAR with 40 fibers. While it has a lower number of fibers than SemanticKITTI, they have a similar vertical angular resolution (0.4° for SemanticKITTI and 0.33° for SemanticPOSS).

\begin{table*}[t]
\scriptsize
    \centering
    \begin{tabular}{c|c||c|c||c|c||c||c}
    Method & Input type & mIoU$_{\mathcal{L}_{SK}}^{SK}$ &mIoU$_{\mathcal{L}_{SK}}^{SK32}$ & mIoU$_{\mathcal{L}_{SK\cap PS}}^{P64}$ & mIoU$_{\mathcal{L}_{SK\cap PS}}^{PFF}$  & mIoU$_{\mathcal{L}_{SK\cap SP}}^{SP}$ & mIoU$_{\mathcal{L}_{SK\cap NS}}^{NS}$\\
    &&&&&&&\\[-0.7em] \hline
    CENet \cite{9859693} & Range-based & 58.8 & 39.1 & 13.3 & 4.9& 27.9 & 5.0  \\ \hline
    Helix4D \cite{loiseau22online} & Point-based \& 4D & 60.0 & 53.2 & 27.7 & 14.2& 36.0 & 34.3  \\ \hline
    KPConv \cite{thomas2019KPConv}& Point-based &58.3& 52.7 & 32.7 & 21.1& 39.1 & \textbf{46.7} \\ \hline
    SRUNet \cite{mink}& Voxel-based&58.6 & 54.0 & \textbf{44.2} & \textbf{22.2}& 45.3 & 42.7   \\ \hline
    SPVCNN \cite{spvnas}& Voxel \& point-based&\textbf{62.3} & \textbf{57.4}& 40.2 & 19.4 & \textbf{45.4} & 45.1  \\\hline
    C3D \cite{Zhou2020Cylinder3DAE}& Cylindrical voxel-based& 60.7& 53.1 & 18.4 & 6.5& 41.9 & 32.7 
    \end{tabular}
    \caption{Generalization benchmark when trained on SemanticKITTI and tested on SemanticKITTI (SK), SemanticKITTI32 (SK32), Panda64 (P64), PandaFF (PFF), SemanticPOSS (PS), and nuScenes (NS).}
    \label{tab:sensor_shift}
\end{table*}

Due to the characteristics of the sensor of Panda64 and SemanticPOSS, we can assume that the sensor shift is minimal, thus they are used to measure scene shift and appearance shift.

nuScenes is acquired by a 360-degree rotating LiDAR with 32 fibers in a urban context. It displays many dynamic objects alongside sensor shift, scene shift, and appearance shift. 

A summary can be found in \autoref{tab:sensor_info}.

\subsubsection{Mean intersection over union}

The evaluation of the transferability of a model from one dataset to another is not straightforward because their label sets are all different. This difference in labelling, coming from the scene shift, and preferences of the authors, makes the comparison more difficult than mIoU computations on the same label set.

To tackle these labeling differences, we realign the label set thanks to a label remapping inspired by \cite{https://doi.org/10.48550/arxiv.2202.06884}. Using similar coarse labels, we group together the labels that do not have a one-to-one mapping from one dataset to another while keeping the labels that do. This coarse-and-fine label set is different for each pair of datasets to measure more precisely the quality of the scene understanding. 

This results in the definition of several label sets, from which we detail the two most important: $\mathcal{L}_{SK\cap SP}=$ \{person, rider, bike, car, ground, trunk, vegetation, traffic-sign, pole, building, fence\} between SemanticKITTI (SK) and SemanticPOSS (SP) and $\mathcal{L}_{SK\cap NS}=$ \{motorcycle, bicycle, person, road, sidewalk, other ground, manmade, vegetation, car, terrain\} between SemanticKITTI (SK) and nuScenes (NS). Details of the mapping from the original label sets to these coarse-and-fine label sets can be found in the Supplementary Material.

In order to clarify which label sets were used in each table, we introduce a specific notation for the mIoU: mIoU$^{target}_{\mathrm{label set}}$. For instance, when we evaluate results on SemanticPOSS (SP) with the label set $\mathcal{L}_{SK\cap SP}$ while the source training set is SemanticKITTI (SK), we compute the mIoU$_{\mathcal{L}_{SK\cap SP}}^{SP}$.

Although these label sets are used for evaluation, they were not used for training purposes. Every method was trained on the regular label set of the source data.

When we evaluate source-to-source segmentation, which means evaluating a model on a validation set that presents no shift from the training set, we compute the results on the original label set.

\subsubsection{Methods}

The methods for this benchmark were chosen based on three factors: availability of the source code and the pretrained models, performances on the SemanticKITTI and nuScenes semantic segmentation benchmarks, and representation of the various LiDAR semantic segmentation paradigms (range based, voxel based, point based, and sequence based). 

As such, we selected 6 different architectures: CENet \cite{9859693}, Helix4D \cite{loiseau22online}, KPConv \cite{thomas2019KPConv}, SRUNet \cite{mink}, SPVCNN \cite{spvnas}, and Cylinder3D \cite{Zhou2020Cylinder3DAE}. 

While this benchmark is not fully exhaustive, we believe it is complete enough to propose an overview of the current generalization performances. 

All these architectures were retrained to omit the reflectivity channel, the details of the training parameters can be found in the Supplementary Material. They were selected by using the official implementations and proposed parameters. We removed the reflectivity channel to improve the generalization performance. Details of the proof can be found in \autoref{sec:r}.

\subsection{Generalization benchmark}
\label{sec:benchmark}
Our proposed generalization benchmark can be found in \autoref{tab:sensor_shift}. We show the mIoU over the 6 generalization cases and on the source-to-source case, here evaluated on Sequence 08 of SemanticKITTI. The details of the per-category IoUs can be found in the Supplementary Material. The most obvious observation that can be made is that high source-to-source segmentation performances do not translate automatically into high generalization performances.

As all methods are trained without reflectivity, they have to rely on geometry only for learning. As such the first experiment (SemanticKITTI to SemanticKITTI32) shows the robustness against the sparsity of the scene. It highlights a very large sensitivity of the sparsity for the range-based method (CENet). Other methods are more robust but are still displaying a significant drop in performance (between -5\% and -7\%).

Panda64 and SemanticPOSS help us measure robustness to scene and appearance shifts. First, we see that performances can change drastically from one dataset to another, especially for Cylinder3D which ranks third on SemanticPOSS, but fifth on Panda64. However, we observe that overall voxel-based methods (SRUNet and SPVCNN) are the most robust, and point-based (KPConv and Helix4D) are right behind.

To further investigate the robustness to sensor shift, we put ourselves in a more difficult case with PandaFF. As the LiDAR is a solid-state LiDAR, the scan topology differs from the typical scans found in the other datasets (only 60 degrees front facing vs 360 degrees for other datasets). Furthermore, the range of the scan is much bigger (150m vs 50m for SemanticKITTI), which can cause issues for hyperparameter selection. We observe that no architecture reaches satisfying results. Nonetheless, KPConv, SRUNet, and SPVCNN are much more robust than other methods. Coupled with the results on Panda64, which displays the exact same scene with a different sensor, we can see that KPConv displays a better robustness to sensors than SRUNet and SPVCNN as it drops less performance from one modality to another. 

Finally, results on nuScenes re-highlight some trends observed previsously. We especially observe the robustness of the voxel-based methods and of KPConv to scene and sensor shifts. We see that the ranking of the methods on nuScenes cannot be predicted by only examining the rankings for SemanticPOSS and SemanticKITTI32, even if together they represent the same shifts as nuScenes.

To conclude, we can see that CENet is overall the least satisfying methods for generalization. Cylinder3D displays a large sensitivity to sensor shift, but can otherwise reach decent performances. Helix4D has more difficulties than KPConv against sensor and scene shifts. Finally, KPConv, SRUNet, and SPVCNN are the most robust methods.

It is interesting to note that while we observe some trends on the generalization abilities of some methods, none of them systematically outperform every other method.

\subsection{Effect of reflectivity on generalization}
\label{sec:r}

As mentioned in the introduction, although reflectivity is a very efficient feature for source-to-source semantic segmentation, we believe it is detrimental to domain generalization and very sensitive to sensor and even firmware changes (e.g., Velodyne HDL32 firmware change\footnote{\href{https://velodynelidar.com/wp-content/uploads/2019/08/63-9114-rev-A-HDL32E-HDL-32E-Software-Version-V2.0-Manual.pdf }{HDL32E Software Manual}}). We demonstrate this in \autoref{tab:DG_R}. We have found a large improvement for generalization when the model does not use reflectivity. In return, we observe a decrease in source-to-source segmentation performances. 

With-reflectivity models were retrieved from the official implementation of the models when they were available. Otherwise we retrained them. We also retrained all reflectivity-less models following the same data augmentation patterns, using the official repositories and their proposed hyperparameters. As mentioned before, details can be found in the Supplementary Material. 

In order to account for the difference in the reflectivity range from SemanticKITTI to SemanticPOSS, the reflectivity was rescaled which is a first step of domain adaptation, to achieve better results, for the models with reflectivity. SemanticPOSS was used for the reflectivity analysis due to the similarity in scan topology with SemanticKITTI and because it follows a more similar reflectivity distribution.

\begin{table}[!h]
\scriptsize
    \centering
    \begin{tabular}{c|c||c||c}
       \multicolumn{1}{c|}{}&Reflectivity&mIoU$_{\mathcal{L}_{SK}}^{SK}$ & mIoU$_{\mathcal{L}_{SK\cap SP}}^{SP}$\\
        &&&\\[-1em] \cline{2-4}
        \multirow{2}{*}{CENet$\dagger$} &yes & \textbf{61.4} &27.5\\\cline{2-4}
        & no&58.8&\textbf{27.9}\\ \hline\hline
        \multirow{2}{*}{Helix4D$\dagger$} &yes & \textbf{63.1} & 35.0 \\\cline{2-4}
        & no& 60.0 & \textbf{36.0}\\ \hline\hline
        \multirow{2}{*}{KPConv*} & yes & \textbf{59.9} &33.1 \\\cline{2-4}
        & no & 58.3 &\textbf{39.1}\\\hline\hline
        \multirow{2}{*}{SRUNet*} &yes & \textbf{61.9} &45.2\\\cline{2-4}
        & no&58.6 &\textbf{45.3}\\ \hline\hline
        \multirow{2}{*}{SPVCNN$\dagger$} &yes & \textbf{63.8} & 36.9\\\cline{2-4}
        & no& 62.3 &\textbf{45.4} \\ \hline\hline
        \multirow{2}{*}{C3D$\dagger$} & yes& \textbf{66.9} & 33.8 \\ \cline{2-4}
       & no& 60.7&\textbf{41.9} 
    \end{tabular}
    \caption{Study of the effect of using reflectivity as input for the neural networks, using SemanticKITTI (SK) for training and SemanticPOSS (SP) for testing. $\dagger$ means that with-reflectivity trained models were extracted from the official repository, * means it was retrained.}
    \label{tab:DG_R}
\end{table}

While source-to-source performances should not be degraded when performing generalization, when building the benchmark, we wanted to focus on generalization performances and thus we wanted to put each method in a favorable light to show an upper bound on their naive generalization methods. This is why the benchmark was built without reflectivity.

\section{Sequence-based semantic segmentation}
\label{sec:semantic}
\subsection{Motivation}

Although autonomous driving 3D semantic segmentation has obtained satisfying results, we highlighted before that domain generalization could be improved and that the segmentation models are sensitive to domain shifts. To counter this, we propose a new approach to semantic segmentation, called 3DLabelProp (\autoref{fig:fastsegnet}). It is a four-step method designed to achieve natively better resilience to domain shifts. 

It works on accumulated point clouds to create a partial independence from the acquisition sensors and leverages the sequentiality of the data through its label propagation module. 

The label propagation module geometrically propagates a part of the labels inferred in the past to the current and newly acquired point cloud. Because it is purely geometric, it is a very fast module with very high accuracy.

Because of the size of accumulated point clouds, it is tedious and slow to apply deep learning methods to it. To speed up the deep learning process while retaining the high-quality geometric information of the accumulated point cloud, we clusterize it to generate smaller point clouds. These small clusters can then be fed to a high-performance neural network while working at a decent speed thanks to the small size of these extracted point clouds. We chose KPConv as our deep learning model because of its high performance on accumulated point clouds and decent generalization performances.

We detail each block of 3DLabelProp in the following subsections.

\subsection{Point cloud accumulation}

To accumulate consecutive LiDAR frames, we apply the state-of-the-art CT-ICP LiDAR odometry \cite{9811849} without loop closure. Once we have the relative pose, we align all the point clouds on the world coordinate. To process the current frame, we accumulate the last 20 frames with it and perform grid subsampling to reduce the memory footprint without altering the geometry. We also remove points too far from the center of the current frames because they add no geometric neighborhoods to the relevant points while having a high memory footprint. 
\begin{figure}[!h]
        \centering
        \subfloat[Propagation module]{
            \includegraphics[width=.45\linewidth]{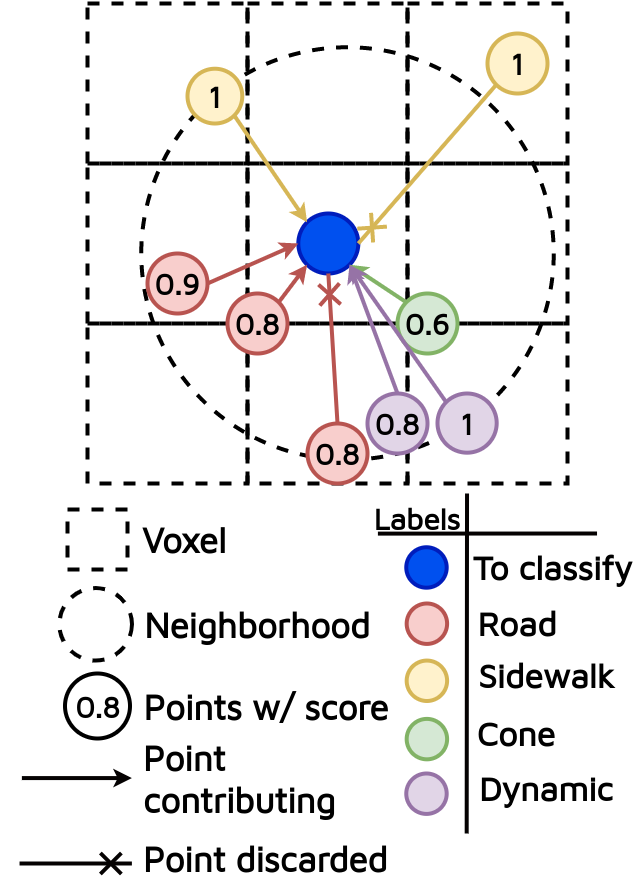}
        }
        \subfloat[Clusterization module]{
            \includegraphics[width=.45\linewidth]{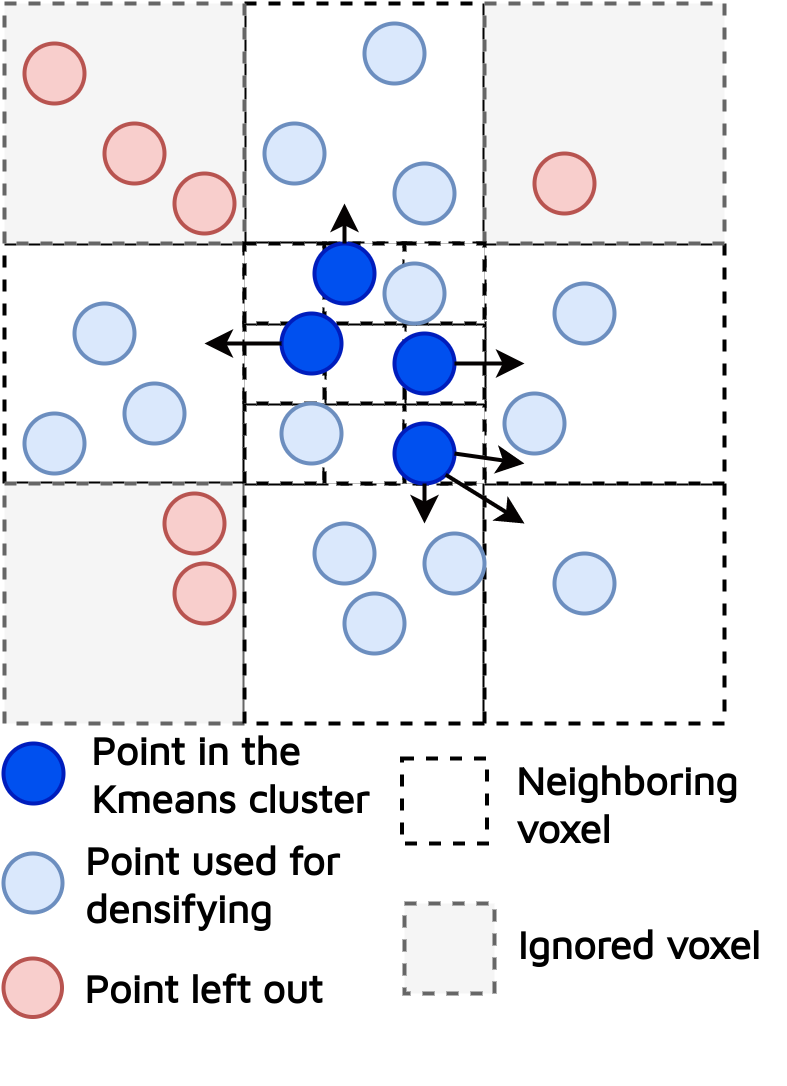}
        } 
    \caption{Illustration of the geometric modules of 3DLabelProp.}
    \label{fig:cluster_and_prop}
\end{figure}
\subsection{Label propagation}
\begin{table*}[h]
\scriptsize
    \centering
    \begin{tabular}{c|c||c|c||c|c||c||c}
    Method & Input type & mIoU$_{\mathcal{L}_{SK}}^{SK}$ &mIoU$_{\mathcal{L}_{SK}}^{SK32}$ & mIoU$_{\mathcal{L}_{SK\cap PS}}^{P64}$ & mIoU$_{\mathcal{L}_{SK\cap PS}}^{PFF}$  & mIoU$_{\mathcal{L}_{SK\cap SP}}^{SP}$ & mIoU$_{\mathcal{L}_{SK\cap NS}}^{NS}$\\
    &&&&&&&\\[-0.7em] \hline
    KPConv \cite{thomas2019KPConv}& Point-based &58.3& 52.7 & 32.7 & 21.1& 39.1 & \textbf{46.7} \\ \hline
    SPVCNN \cite{spvnas}& Voxel \& point-based&\textbf{62.3} & \textbf{57.4}& 40.2 & 19.4 & 45.4 & 45.1  \\\hline
    C3D \cite{Zhou2020Cylinder3DAE}& Cylindrical voxel-based& 60.7& 53.1 & 18.4 & 6.5& 41.9 & 32.7 \\ \hline\hline
    3DLabelProp (Ours) & Point-based \& 4D& 60.8& 56.7 & \textbf{56.0} & \textbf{55.2}& \textbf{50.4} & 44.4 
    \end{tabular}
    \caption{Generalization results when trained on SemanticKITTI and tested on SemanticKITTI (SK), SemanticKITTI32 (SK32), Panda64 (P64), PandaFF (PFF), SemanticPOSS (PS), and nuScenes (NS).}
    \label{tab:DG_from_SK}
\end{table*}
Contrary to offline semantic segmentation, sequence-based semantic segmentation incorporates incremental information in the accumulated point cloud and retains past predictions. Although single scan methods completely discard it, in 3DLabelProp, we leverage this past information to propagate information geometrically in the newly registered scan.

Defining $L = (l_i)_{i \in \{0,K\}}$ as the semantic label set, we split it into two subsets, $D = (d_i)_{i \in \{0,K_d\}}$ and $S = (s_i)_{i \in \{K_d,K\}}$ dynamic and static labels, respectively, such that $L=D\cup S$. Static labels are defined as points that cannot move in the world coordinate, from one scan to another. On the contrary, a dynamic object \textit{can} move from one scan to another. The ground and buildings are examples of static labels, whereas cars and pedestrians are examples of the dynamic labels.

As defined, objects that are part of static labels are expected not to move. Furthermore, two very close LiDAR points in the world coordinates can be assumed to be part of the same static object. As such, we can \textit{propagate} past inferences to present points if they are close enough that they are expected to be part of the same object. To achieve this, we process it in a two-step pipeline. First, we gather neighborhoods for each point of the newly acquired scan, and then, we assign it a propagation score representing confidence in a geometrical prediction.

To extract neighbors, we assess a sphere centered on the current point processed and gathering all points belonging to the past accumulation inside. This neighbor search is sped up by the grid subampling beforehand and the ability to leverage these pre-computed voxels. Then, based on the label and its associated score of each neighbor, we proceed to voting to extract the predicted label. Neighbors that have a distance-weighted confidence too low are discarded as they are considered unreliable.

If the majority vote is a dynamic label then we assign no value. Otherwise, we assign the majority label and the weighted average of the confidence of the neighbors belonging to this label as the new score. The process is illustrated in \autoref{fig:cluster_and_prop}.

We introduce $ \mathcal{P}_{acc}\in 	\mathbb{R}^{N \times 3}$ , the accumulated point clouds made of the past sequence; $Y_{acc}\in \{0,K\}^N$ its associated inferred labels, and $\mathcal{C}_{acc} = (c_{acc,i}\times e_{y_i})_{i\in\{1,N\}}$ the one-hot encoded confidence vector associated with each label, $c_{acc,i}$ the confidence score, and $e_{y_i}$ the one-hot vector with the one at the $y_i$th position. Similarly, we introduce  $ \mathcal{P}_{curr}\in 	\mathbb{R}^{M \times 3}$, $Y_{curr}$, and $\mathcal{C}_{curr}$ which are the associated values for the current LiDAR frame. Because it is not yet processed, the labels are initialized at -1, and the associated confidence is zero. We then proceed to the propagation:

 \begin{center}
    $\forall p_{curr,i}, i \in \{1,M\};$

\begin{equation}
c_{curr,i} = \sum_{p_{acc,n} \in \mathcal{N}(p_{curr,i})}  w(i,n)    \mathds{1}_{w(i,n)>0.5};
\end{equation}
    \end{center}

\noindent $\mathrm{with}\; w(i,n) = d(p_{acc,n},p_{curr,i})\times c_{acc,i} \;
\mathrm{and}\; d(p,q) = e^{-\frac{||p-q||^2}{\sigma^2}}$, where $\sigma$ being the bandwidth hyperparameter, giving the range of the neighbor search. We add a cut-off function in which points with too low confidence relative to the position of the current points are discarded. $\mathcal{N}(p_{curr,i})$ is the neighborhood of $p_{curr,i}$.  Because the past sequence $\mathcal{P}_{acc}$ is grid-subsampled to keep the memory consumption low, we reuse this voxelization to perform a radius-based neighbor search. Because we know the bandwidth of the label propagation, we can restrict our search to the voxels contained in this sphere, hence speeding up the process.
Then we have the following:

\begin{equation}
        y_{curr,i} = 
\begin{cases}
    \argmax(c_{curr,i}), \ \text{if } &\argmax(c_{curr,i}) > K_d \\
-1,              & \text{otherwise}
\end{cases}
\end{equation}

with $\argmax(c_{curr,i}) > K_d$, meaning it is a static label ($K_d$ is the number of dynamic labels).

When fed with past ground truth point clouds, this step reaches $95\%$ accuracy over $80\%$ of the static points. Only $15\% $ of the dynamic points receives a wrongful prediction, making it a robust step.

\subsection{Clusterization}

To efficiently process a large point cloud with KPConv, it needs to be split into smaller point clouds. While the original authors created sub-point clouds by extracting spheres inside the original one, the number of spheres necessary to cover a whole accumulated point cloud is prominent, resulting in a large amount of overlap. To counter this, we propose using a clustering algorithm, covering the point cloud more efficiently and with minimal overlap.

Because we do not need to infer the whole accumulated point clouds but only the remaining points after the label propagation step, we apply the clustering algorithms only on $p_{curr,i}$, where $ y_{curr,i} = -1$, which results in fewer points than the original point clouds and allows for a sophisticated clustering algorithm to run at a high speed. As such, we use the K-means algorithm to create $N_c$ clusters. 

This step creates very sparse clusters that are then completed with points from the accumulated point cloud to create dense neighborhoods for those few remaining un-inferred points. In order to do this, we voxelize the accumulated point cloud, and we fetch each point that belongs to either a current voxel occupied by points of the cluster or points belonging to neighbors of a voxel occupied by points of the cluster. 

Not all the voxels touching the current one are neighbor voxels. Rather, they are computed based on where inside the current voxel the points of the cluster are, precisely. To do this, we divide the voxel into 27 sub-voxels (3x3x3) and assign to each sub-voxel a specific neighborhood. For instance in 2D we divide each pixel in 9 sub-pixel (3x3), and for the middle-left pixel the only neighbor is the left-neighbor. The process is illustrated in 2D in \autoref{fig:cluster_and_prop}.

These completed clusters are the input of our KPConv module.

\subsection{Deep learning}

Finally, we feed the clusters to KPConv, which performs very efficiently on dense point clouds. Because the clusters are relatively small, we reach a higher inference speed than typical KPConv use cases. Because the clusters have a small overlap resulting from cluster completion, some points can undergo several inferences. To have the final predicted label, we average all confidence scores and take the highest value. This average score is the newly assigned confidence.

The resulting pipeline allows us to process accumulated point clouds at a higher pace than typical approaches, achieving very good generalization performance.

The details of the KPConv model used in our pipeline can be found in the supplementary materials.

\subsection{Final prediction}

The last step is to fuse predictions from the propagation module and the deep learning module as some points undergo predictions from both modules. In order to achieve this, we weight the predictions of the label propagation module by w1 and of the deep learning module by w2. This novel score is then used to infer the final label predicted. In pratice, as KPConv is much more reliable than the label propagation module we use w1 = 0 and w2 = 1.

\begin{table*}[h]
\scriptsize
    \centering
    \begin{tabular}{c|c||c||c|c||c|c||c}
    Method & Input type & mIoU$_{\mathcal{L}_{NS}}^{NS}$ & mIoU$_{\mathcal{L}_{NS\cap SK}}^{SK}$ &mIoU$_{\mathcal{L}_{NS\cap SK}}^{SK32}$ & mIoU$_{\mathcal{L}_{NS\cap PS}}^{P64}$ & mIoU$_{\mathcal{L}_{NS\cap PS}}^{PFF}$  & mIoU$_{\mathcal{L}_{NS\cap SP}}^{SP}$ \\
    &&&&&&&\\[-0.7em] \hline
    KPConv \cite{thomas2019KPConv}& Point-based &63.1&44.9&50.6 &25.0 & 16.9  & 60.7\\ \hline
    SPVCNN \cite{spvnas}& Voxel \& point-based& 67.2& 49.4 & 53.2 & 43.7 & 11.1 & \textbf{64.8} \\\hline
    C3D \cite{Zhou2020Cylinder3DAE}& Cylindrical voxel-based&70.2& 31.7&46.1 & 15.8 &4.7 &42.8 \\ \hline\hline
    3DLabelProp (Ours) & Point-based \& 4D&\textbf{71.0}&\textbf{60.5}& \textbf{62.5}& \textbf{65.4} & \textbf{66.7} &64.3 
    \end{tabular}
    \caption{Generalization results when trained on nuScenes and tested on nuScenes (NS), SemanticKITTI (SK), SemanticKITTI32 (SK32), Panda64 (P64), PandaFF (PFF), and SemanticPOSS (PS).}
    \label{tab:DG_from_NS}
\end{table*}

In addition to reproducing the same experiments as the benchmark, we expanded it by training on nuScenes and testing on the same datasets in \autoref{tab:DG_from_NS}.
\section{Results}
\label{sec:results}
In this section, we discuss and study 3DLabelProp to understand the impact of the various implementation choices on the generalization performance. Rather than copying the full benchmark from \autoref{sec:benchmark}, we only write results from Cylinder3D, SPVCNN, and KPConv for comparison with our method. This choice was made as KPConv is the baseline for our model, Cylinder3D is the standing state-of-the-art method for source-to-source segmentation (when reflectivity is available), and SPVCNN showed the overall best generalization performances.

\subsection{Domain generalization from SemanticKITTI}

We detail the results from 3DLabelProp in \autoref{tab:DG_from_SK}. First, we observe that 3DLabelProp reaches satisfying source-to-source segmentation results, outperforming KPConv by more than 2\%. 

We built 3DLabelProp to decrease sensitivity to domain shift and precisely to sensor shift by working on the accumulated point cloud. 3DLabelProp displays the smallest decrease in performance from SemanticKITTI to SemanticKITTI32 (-4\%) and between Panda64 and PandaFF (-0.8\%). It is also the only method reaching satisfying performance on PandaFF (+33\% from the second best).

Finally, on generalization performances, 3DLabelProp reaches the best performances for Panda64 (+12\% from second best) and for SemanticPOSS (+5\% from second best) and is the third best (-2.3\%) on nuScenes. As we demonstrated the relatively low sensitivity from sensor shift for 3DLabelProp, nuScenes' results stem from scene shift. In particular, when studying per-class results in the Supplementary Material, we observe that the performances of KPConv on pedestrians explain its results.

Overall, 3DLabelProp is the state of the art for generalization regarding robustness and performance.

\subsection{Domain generalization from nuScenes}

For the state-of-the-art methods, we see similar patterns when trained with nuScenes and trained with SemanticKITTI. We note the sensitivity to domain shifts, as methods systematically perform better on SemanticKITTI32 than SemanticKITTI.

For source-to-source segmentation, we see that 3DLabelProp reaches state-of-the-art performance, showing that working on a denser representation is helpful, especially for low resolution LiDARs.  Furthermore, we see patterns of results similar to results from \autoref{tab:DG_from_SK}. We see relatively low sensitivity to sensor shift as there is a very limited performance change from SemanticKITTI32 to SemanticKITTI (-2\%), and the lowest change in performance between Panda64 and PandaFF (-1\%). It is also on par with SPVCNN for SemanticPOSS (-0.5\%) and outperforming it for Panda64 (+21.7\%).


\subsection{Related work comparison}

Similarly to Complete \& Label (C\&L) \cite{9578920}, we believe that a dense representation of point clouds could reduce the domain gap. The way to achieve this representation is different. We accumulate our point clouds based on geometrical information, whereas they rely on learning of a completion network. We ensure the preservation of geometric properties scenes without additional hyperparameters. We compare our generalization results with theirs in \autoref{tab:cl}. Furthermore, we compute the relative decrease from source-to-source to generalization to highlight the effectiveness of our method compared to theirs (which is Unsupervised Domain Adaptation, UDA, training using target data information without target labels, whereas we use no information about the target set). We are much more accurate from source-to-source and are either on par or quite better regarding relative decrease. We believe it stems from their difficulty in effectively training a completion model that generalizes well, whereas we use only the geometry for domain alignment.

\begin{table}[h]
\scriptsize
    \centering
    \begin{tabular}{c|c|c|c|c|c|c}
     \multicolumn{1}{c|}{}&\multicolumn{3}{c|}{SK $\rightarrow$ NS}&\multicolumn{3}{|c}{NS $\rightarrow$ SK} \\ \hline
        Method & SK & NS & \% drop & NS & SK &  \% drop \\\hline 
        C\&L \cite{9578920} & 50.2 & 31.6 & -37\% & 54.4 & 33.7 & -38\%\\\hline 
        3DLabelProp & 69.0 & 42.7 & -38\% & 66.5 & 50.5 & -24\%
    \end{tabular}\par
    
    \caption{Comparison with C\&L. All results were computed on the C\&L label set.}
    \label{tab:cl}
\end{table}

\subsection{Ablation study}
\label{sec:as}

\begin{table}[!h]
\scriptsize
    \centering
    \begin{tabular}{c|c|c|c|c}
       &mIoU$_{\mathcal{L}_{SK}}^{SK}$  & mIoU$_{\mathcal{L}_{SK\cap SP}}^{SP}$ & mIoU$_{\mathcal{L}_{SK\cap NS}}^{NS}$ &FPS\\
    &&&&\\[-0.7em] \hline
     \specialcell{KPConv \\ w/ reflectivity} & 59.9 & 33.1&47.6 & 0.2\\\hline
     \specialcell{KPConv } & 58.3  & 39.1 &46.7 & 0.2\\\hline
     \specialcell{KPConv \\multiframe} & 53.0 &47.2& 44.2 & 0.05 \\\hline
     \specialcell{3DLabelProp \\ (Ours)} & 60.8 & 50.4 & 44.4 & 1.0 
    \end{tabular}
    \caption{Ablation study on 3DLabelProp}
    \label{tab:ab_KP}
\end{table}

In the ablation study, we examined the impact of several design choices in the conception of 3DLabelProp. All variants (including 3DLabelProp) use the same model design, which is the one provided by the authors of KPConv. Training parameters differ for 3DLabelProp compared to the standard KPConv, due to the format of the input point cloud, which is a sphere for KPConv, and a cluster of unknown shape for 3DLabelProp. For KPConv multiframe we used the past 10 scans for accumulation.

First, we again note the results of KPConv with or without the reflectivity channel. Dropping the reflectivity channel triggers a huge gain in generalization from SemanticKITTI to SemanticPOSS for a very small loss from SemanticKITTI to nuScenes. The second step is computing KPConv on an accumulated frame, which also helps improve the generalization performances. Accumulating frames in SemanticPOSS and SemanticKITTI accentuates their geometric similarity, which results in extremely dense point clouds.

Using the full 3DLabelProp has two advantages: it keeps on improving the overall generalization performance when compared with other KPConv approaches and the various geometric modules help speed up the inference to better reach decent performances. The speed is computed for the inference on nuScenes.

\subsection{Geometric parameters}
For the propagation and clusterization steps of 3DLabelProp, we introduced 5 parameters: $v_s$, size of the voxels for the grid subsampling; $d_{prop}$, the distance of the propagation, which is linked to the bandwidth parameter $\sigma$; $N_c$, the number of clusters; $V_c$, the size of the voxels for the cluster densification; and $N_f$ the number for frames used in the accumulated point cloud. The relationship between $d_{prop}$ and $\sigma$, the bandwith parameter, is: $\sigma = \frac{d_{prop}}{\sqrt{ln(2)}}$

For all the trainings and all the inferences, the parameters chosen were $v_s$ = 0.05m, $d_{prop}$= 0.30m, $V_c$ = 2m, and $N_c$ = 10 when the training data is SemanticKITTI, and $N_c$ = 20 when the training data is nuScenes. Only $N_c$ changes depending on the training model. We have a larger $N_c$ for nuScenes as the training data is sparser and we want to create more overlap between clusters.

It is important to note that to respect the principle of generalization, these parameters are chosen on the training set and are not modified for all the inferences on all the generalization datasets. The two most important parameters of this geometric part are $d_{prop}$ for the propagation distance and $N_c$ for the number of clusters. We did an ablation study of these parameters in \autoref{tab:new_ab} by training on nuScenes and testing the generalization on SemanticKITTI. We chose nuScenes as the training data due to its smaller training time.

\begin{table}[]
\scriptsize
    \centering
    \begin{tabular}{c|c|c|c|c|c}
         $d_{prop}$ & $N_c$ & $N_f$ & mIoU$_{\mathcal{L}_{NS}}^{NS}$ & mIoU$_{\mathcal{L}_{NS\cap SK}}^{SK}$ & FPS\\
         &&&&\\[-0.7em] \hline\hline
         0.30m&20&20&71.0&60.5&1.2\\\hline\hline
         0.10m&20&20&72.4&60.2&0.6\\ \hline
         0.60m&20&20&69.1&57.2&1.5\\ \hline\hline
         0.30m&5&20&71.3&61.5&1.3\\ \hline
         0.30m&40&20&68.8&59.7&0.9\\\hline\hline
         0.30m&20&5&67.4&60.0&1.5\\\hline
         0.30m&20&10&70.9&60.2&1.5\\ \hline
         0.30m&20&40&67.9&59.5&1.0\\ 
    \end{tabular}
    \caption{Study of the geometric parameters. The first line is the model used in the paper.}
    \label{tab:new_ab}
\end{table}

First, we observe that even with ill-selected parameters ($d_{prop}$=0.60m), 3DLabelProp still achieves state-of-the-art generalization from nuScenes to SemanticKITTI. Then, we see the robustness, performance-wise, to the parameter selection, as there is only a difference of 4\% between the best and worst performances. 

The largest impact is on the process time. We see that $d_{prop}$ is the main source of speed variation. We selected $d_{prop}$=0.30m as a trade-off between speed and accuracy, as the increase in performance between 0.30m and 0.10m is minimal but the speed drop is quite large.

The cluster parameter is more linked to memory consumption. The larger the number of clusters, the less memory consumption, up until a certain point where too many clusters create a lot of overlap and re-increase the memory consumption. It is not reported here as it has a small impact on nuScenes, but can cause issue in denser point cloud such as SemanticKITTI, which explains our choice of $N_c$=20 rather than $N_c$=5. 

\section{Conclusion}
\label{sec:ccl}

We have proposed the first 3D autonomous driving semantic segmentation domain generalization benchmark, highlighting the necessary work to be done in this area. We specifically highlighted how overfitting  models on source-to-source segmentation is detrimental to domain generalization performance, whether done through architecture design or feature vector selection.

To fill the gap in the literature, we have proposed the first method designed for this task. Using two different source datasets and over 5 other target datasets we demonstrated 3DLabelProp is the most balanced and accurate generalization method.

{\small
\bibliographystyle{ieee_fullname}
\bibliography{egbib}

\begin{thebibliography}{10}\itemsep=-1pt

\bibitem{aygun20214d}
Mehmet Aygun, Aljosa Osep, Mark Weber, Maxim Maximov, Cyrill Stachniss, Jens
  Behley, and Laura Leal-Taixe.
\newblock 4d panoptic lidar segmentation.
\newblock In {\em Proceedings of the IEEE/CVF Conference on Computer Vision and
  Pattern Recognition (CVPR)}, pages 5527--5537, June 2021.

\bibitem{behley2019iccv}
Jens Behley, Martin Garbade, Andres Milioto, Jan Quenzel, Sven Behnke, Cyrill
  Stachniss, and Jürgen Gall.
\newblock Semantickitti: A dataset for semantic scene understanding of lidar
  sequences.
\newblock In {\em 2019 IEEE/CVF International Conference on Computer Vision
  (ICCV)}, pages 9296--9306, 2019.

\bibitem{nuscenes2019}
Holger Caesar, Varun Bankiti, Alex~H. Lang, Sourabh Vora, Venice~Erin Liong,
  Qiang Xu, Anush Krishnan, Yu Pan, Giancarlo Baldan, and Oscar Beijbom.
\newblock nuscenes: A multimodal dataset for autonomous driving.
\newblock In {\em 2020 IEEE/CVF Conference on Computer Vision and Pattern
  Recognition (CVPR)}, pages 11618--11628, 2020.

\bibitem{8953760}
Fabio~M. Carlucci, Antonio D'Innocente, Silvia Bucci, Barbara Caputo, and
  Tatiana Tommasi.
\newblock Domain generalization by solving jigsaw puzzles.
\newblock In {\em 2019 IEEE/CVF Conference on Computer Vision and Pattern
  Recognition (CVPR)}, pages 2224--2233, 2019.

\bibitem{9859693}
Hui–Xian Cheng, Xian–Feng Han, and Guo–Qiang Xiao.
\newblock Cenet: Toward concise and efficient lidar semantic segmentation for
  autonomous driving.
\newblock In {\em 2022 IEEE International Conference on Multimedia and Expo
  (ICME)}, pages 01--06, 2022.

\bibitem{mink}
Christopher Choy, JunYoung Gwak, and Silvio Savarese.
\newblock 4d spatio-temporal convnets: Minkowski convolutional neural networks.
\newblock In {\em 2019 IEEE/CVF Conference on Computer Vision and Pattern
  Recognition (CVPR)}, pages 3070--3079, 2019.

\bibitem{9811849}
Pierre Dellenbach, Jean-Emmanuel Deschaud, Bastien Jacquet, and François
  Goulette.
\newblock Ct-icp: Real-time elastic lidar odometry with loop closure.
\newblock In {\em 2022 International Conference on Robotics and Automation
  (ICRA)}, pages 5580--5586, 2022.

\bibitem{deschaud2021kitticarla}
Jean-Emmanuel {Deschaud}.
\newblock {KITTI-CARLA: a KITTI-like dataset generated by CARLA Simulator}.
\newblock {\em arXiv e-prints}, 2021.

\bibitem{fan2021pstnet}
Hehe Fan, Xin Yu, Yuhang Ding, Yi Yang, and Mohan Kankanhalli.
\newblock Pstnet: Point spatio-temporal convolution on point cloud sequences.
\newblock In {\em International Conference on Learning Representations}, 2021.

\bibitem{10.1007/978-3-031-19809-0_16}
H.~Huang C. Chen~Y. Fang.
\newblock Manifold adversarial learning for cross-domain 3d shape
  representation.
\newblock In {\em ECCV 2022}, 2022.

\bibitem{9710159}
D. Hendrycks, S. Basart, N. Mu, S. Kadavath, F. Wang, E. Dorundo, R. Desai, T.
  Zhu, S. Parajuli, M. Guo, D. Song, J. Steinhardt, and J. Gilmer.
\newblock The many faces of robustness: A critical analysis of
  out-of-distribution generalization.
\newblock In {\em 2021 IEEE/CVF International Conference on Computer Vision
  (ICCV)}, pages 8320--8329, Los Alamitos, CA, USA, oct 2021. IEEE Computer
  Society.

\bibitem{panda}
Scale.AI Hesai.
\newblock {Pandaset open datasets}.
\newblock https://scale.com/open-datasets/pandaset, 2020.
\newblock [Online; accessed 18-February-2022].

\bibitem{horache2021mssvconv}
S. Horache, J. Deschaud, and F. Goulette.
\newblock 3d point cloud registration with multi-scale architecture and
  unsupervised transfer learning.
\newblock In {\em 2021 International Conference on 3D Vision (3DV)}, pages
  1351--1361, Los Alamitos, CA, USA, dec 2021. IEEE Computer Society.

\bibitem{8237429}
Xun Huang and Serge Belongie.
\newblock Arbitrary style transfer in real-time with adaptive instance
  normalization.
\newblock In {\em 2017 IEEE International Conference on Computer Vision
  (ICCV)}, pages 1510--1519, 2017.

\bibitem{9513542}
Xin Jin, Cuiling Lan, Wenjun Zeng, and Zhibo Chen.
\newblock Style normalization and restitution for domain generalization and
  adaptation.
\newblock {\em IEEE Transactions on Multimedia}, 24:3636--3651, 2022.

\bibitem{kreuzberg2022stop}
Lars Kreuzberg, Idil~Esen Zulfikar, Sabarinath Mahadevan, Francis Engelmann,
  and Bastian Leibe.
\newblock 4d-stop: Panoptic segmentation of 4d lidar using spatio-temporal
  object proposal generation and aggregation.
\newblock In {\em European Conference on Computer Vision Workshop}, 2022.

\bibitem{9341508}
Ferdinand Langer, Andres Milioto, Alexandre Haag, Jens Behley, and Cyrill
  Stachniss.
\newblock Domain transfer for semantic segmentation of lidar data using deep
  neural networks.
\newblock In {\em 2020 IEEE/RSJ International Conference on Intelligent Robots
  and Systems (IROS)}, pages 8263--8270, 2020.

\bibitem{9879166}
A. Lehner, S. Gasperini, A. Marcos-Ramiro, M. Schmidt, M. Mahani, N. Navab, B.
  Busam, and F. Tombari.
\newblock 3d-vfield: Adversarial augmentation of point clouds for domain
  generalization in 3d object detection.
\newblock In {\em CVPR}, pages 17274--17283, 2022.

\bibitem{Li2018MLDG}
Da Li, Yongxin Yang, Yi-Zhe Song, and Timothy Hospedales.
\newblock Learning to generalize: Meta-learning for domain generalization.
\newblock In {\em AAAI Conference on Artificial Intelligence}, 2018.

\bibitem{9008109}
Da Li, Jianshu Zhang, Yongxin Yang, Cong Liu, Yi-Zhe Song, and Timothy
  Hospedales.
\newblock Episodic training for domain generalization.
\newblock In {\em 2019 IEEE/CVF International Conference on Computer Vision
  (ICCV)}, pages 1446--1455, 2019.

\bibitem{360}
Yiyi Liao, Jun Xie, and Andreas Geiger.
\newblock {KITTI}-360: A novel dataset and benchmarks for urban scene
  understanding in 2d and 3d.
\newblock {\em Pattern Analysis and Machine Intelligence (PAMI)}, 2022.

\bibitem{9010250}
Xingyu Liu, Mengyuan Yan, and Jeannette Bohg.
\newblock Meteornet: Deep learning on dynamic 3d point cloud sequences.
\newblock In {\em 2019 IEEE/CVF International Conference on Computer Vision
  (ICCV)}, pages 9245--9254, 2019.

\bibitem{loiseau22online}
Romain Loiseau, Mathieu Aubry, and Loic Landrieu.
\newblock Online segmentation of lidar sequences: Dataset and algorithm.
\newblock {\em ECCV}, 2022.

\bibitem{9796597}
Benedikt Mersch, Xieyuanli Chen, Ignacio Vizzo, Lucas Nunes, Jens Behley, and
  Cyrill Stachniss.
\newblock Receding moving object segmentation in 3d lidar data using sparse 4d
  convolutions.
\newblock {\em IEEE Robotics and Automation Letters}, 7(3):7503--7510, 2022.

\bibitem{8967762}
Andres Milioto, Ignacio Vizzo, Jens Behley, and Cyrill Stachniss.
\newblock Rangenet ++: Fast and accurate lidar semantic segmentation.
\newblock In {\em 2019 IEEE/RSJ International Conference on Intelligent Robots
  and Systems (IROS)}, pages 4213--4220, 2019.

\bibitem{pan2018IBN-Net}
Xingang Pan, Ping Luo, Jianping Shi, and Xiaoou Tang.
\newblock Two at once: Enhancing learning and generalization capacities via
  ibn-net.
\newblock In {\em ECCV}, 2018.

\bibitem{poss}
Yancheng Pan, Biao Gao, Jilin Mei, Sibo Geng, Chengkun Li, and Huijing Zhao.
\newblock Semanticposs: A point cloud dataset with large quantity of dynamic
  instances.
\newblock In {\em 2020 IEEE Intelligent Vehicles Symposium (IV)}, pages
  687--693, 2020.

\bibitem{https://doi.org/10.48550/arxiv.2202.06884}
Jules Sanchez, Jean-Emmanuel Deschaud, and François Goulette.
\newblock Cola: Coarse label pre-training for 3d semantic segmentation of
  sparse lidar datasets, 2022.

\bibitem{spvnas}
Haotian Tang, Zhijian Liu, Shengyu Zhao, Yujun Lin, Ji Lin, Hanrui Wang, and
  Song Han.
\newblock Searching efficient 3d architectures with sparse point-voxel
  convolution.
\newblock In Andrea Vedaldi, Horst Bischof, Thomas Brox, and Jan-Michael Frahm,
  editors, {\em Computer Vision -- ECCV 2020}, pages 685--702, Cham, 2020.
  Springer International Publishing.

\bibitem{Tat2018}
Maxim Tatarchenko, Jaesik Park, Vladlen Koltun, and Qian-Yi Zhou.
\newblock Tangent convolutions for dense prediction in 3d.
\newblock In {\em 2018 IEEE/CVF Conference on Computer Vision and Pattern
  Recognition}, pages 3887--3896, 2018.

\bibitem{thomas2019KPConv}
Hugues Thomas, Charles~R. Qi, Jean-Emmanuel Deschaud, Beatriz Marcotegui,
  François Goulette, and Leonidas Guibas.
\newblock Kpconv: Flexible and deformable convolution for point clouds.
\newblock In {\em 2019 IEEE/CVF International Conference on Computer Vision
  (ICCV)}, pages 6410--6419, 2019.

\bibitem{NEURIPS2018_1d94108e}
Riccardo Volpi, Hongseok Namkoong, Ozan Sener, John~C Duchi, Vittorio Murino,
  and Silvio Savarese.
\newblock Generalizing to unseen domains via adversarial data augmentation.
\newblock In S. Bengio, H. Wallach, H. Larochelle, K. Grauman, N. Cesa-Bianchi,
  and R. Garnett, editors, {\em Advances in Neural Information Processing
  Systems}, volume~31. Curran Associates, Inc., 2018.

\bibitem{9577308}
H~.Chao C. Zhangjie Y. Wang~J. Wang and M. Long.
\newblock Metasets: Meta-learning on point sets for generalizable
  representations.
\newblock In {\em 2021 CVPR}, 2021.

\bibitem{9782500}
Jindong Wang, Cuiling Lan, Chang Liu, Yidong Ouyang, Tao Qin, Wang Lu, Yiqiang
  Chen, Wenjun Zeng, and Philip Yu.
\newblock Generalizing to unseen domains: A survey on domain generalization.
\newblock {\em IEEE Transactions on Knowledge and Data Engineering}, pages
  1--1, 2022.

\bibitem{wu2017squeezeseg}
Bichen Wu, Alvin Wan, Xiangyu Yue, and Kurt Keutzer.
\newblock Squeezeseg: Convolutional neural nets with recurrent crf for
  real-time road-object segmentation from 3d lidar point cloud.
\newblock {\em ICRA}, 2018.

\bibitem{xiao2022transfer}
Aoran Xiao, Jiaxing Huang, Dayan Guan, Fangneng Zhan, and Shijian Lu.
\newblock Transfer learning from synthetic to real lidar point cloud for
  semantic segmentation.
\newblock In {\em Proceedings of the AAAI Conference on Artificial
  Intelligence}, volume~36, pages 2795--2803, 2022.

\bibitem{squeezesegv3}
Chenfeng Xu, Bichen Wu, Zining Wang, Wei Zhan, Peter Vajda, Kurt Keutzer, and
  Masayoshi Tomizuka.
\newblock Squeezesegv3: Spatially-adaptive convolution for efficient
  point-cloud segmentation.
\newblock In Andrea Vedaldi, Horst Bischof, Thomas Brox, and Jan-Michael Frahm,
  editors, {\em Computer Vision -- ECCV 2020}, pages 1--19, Cham, 2020.
  Springer International Publishing.

\bibitem{9896915}
Jiabo Xu, Yukun Huang, Zeyun Wan, and Jingbo Wei.
\newblock Glorn: Strong generalization fully convolutional network for
  low-overlap point cloud registration.
\newblock {\em IEEE Transactions on Geoscience and Remote Sensing}, pages 1--1,
  2022.

\bibitem{Xu2021RPVNetAD}
Jianyun Xu, Ruixiang Zhang, Jian Dou, Yushi Zhu, Jie Sun, and Shiliang Pu.
\newblock Rpvnet: A deep and efficient range-point-voxel fusion network for
  lidar point cloud segmentation.
\newblock In {\em Proceedings of the IEEE/CVF International Conference on
  Computer Vision (ICCV)}, pages 16024--16033, October 2021.

\bibitem{10.1007/978-3-030-01237-3_6}
Yifan Xu, Tianqi Fan, Mingye Xu, Long Zeng, and Yu Qiao.
\newblock Spidercnn: Deep learning on point sets with parameterized
  convolutional filters.
\newblock In Vittorio Ferrari, Martial Hebert, Cristian Sminchisescu, and Yair
  Weiss, editors, {\em Computer Vision -- ECCV 2018}, pages 90--105, Cham,
  2018. Springer International Publishing.

\bibitem{randconv2021}
Z. Xu, D. Liu, J. Yang, C. Raffel, and M. Niethammer.
\newblock Robust and generalizable visual representation learning via random
  convolutions.
\newblock In {\em ICLR}, 2021.

\bibitem{yan20222dpass}
Xu Yan, Jiantao Gao, Chaoda Zheng, Chao Zheng, Ruimao Zhang, Shuguang Cui, and
  Zhen Li.
\newblock 2dpass: 2d priors assisted semantic segmentation on lidar point
  clouds.
\newblock In {\em ECCV}, 2022.

\bibitem{9578920}
L. Yi, B. Gong, and T. Funkhouser.
\newblock Complete \& label: A domain adaptation approach to semantic
  segmentation of lidar point clouds.
\newblock In {\em CVPR}, 2021.

\bibitem{Zhang_2020_CVPR}
Yang Zhang, Zixiang Zhou, Philip David, Xiangyu Yue, Zerong Xi, Boqing Gong,
  and Hassan Foroosh.
\newblock Polarnet: An improved grid representation for online lidar point
  clouds semantic segmentation.
\newblock In {\em 2020 IEEE/CVF Conference on Computer Vision and Pattern
  Recognition (CVPR)}, pages 9598--9607, 2020.

\bibitem{Zhao2021ePointDAAE}
Sicheng Zhao, Yezhen Wang, Bo Li, Bichen Wu, Yang Gao, Pengfei Xu, Trevor
  Darrell, and Kurt Keutzer.
\newblock epointda: An end-to-end simulation-to-real domain adaptation
  framework for lidar point cloud segmentation.
\newblock In {\em AAAI}, 2021.

\bibitem{9847099}
Kaiyang Zhou, Ziwei Liu, Yu Qiao, Tao Xiang, and Chen~Change Loy.
\newblock Domain generalization: A survey.
\newblock {\em IEEE Transactions on Pattern Analysis and Machine Intelligence},
  pages 1--20, 2022.

\bibitem{Zhou2020Cylinder3DAE}
Xinge Zhu, Hui Zhou, Tai Wang, Fangzhou Hong, Yuexin Ma, Wei Li, Hongsheng Li,
  and Dahua Lin.
\newblock Cylindrical and asymmetrical 3d convolution networks for lidar
  segmentation.
\newblock In {\em 2021 IEEE/CVF Conference on Computer Vision and Pattern
  Recognition (CVPR)}, pages 9934--9943, 2021.

\end{thebibliography}
}

\appendix
\clearpage
\section{Implementation details}
All results presented in this paper were computed on a GPU Nvidia RTX 3090.

\subsection{State-of-the-art architectures}


 To compute the inference and for re-training, the following open-source codes were used (all found on github), alongside these associated parameters:
\begin{itemize}
    \item KPConv: \href{https://github.com/HuguesTHOMAS/KPConv-PyTorch}{HuguesTHOMAS/KPConv-PyTorch}. Model: KPFCNN; learning rate 1e-2; number of iterations: 500 000; maximum number of points per batch: 13400. Same parameters for nuScenes and SemanticKITTI. The choice of accumulating the 10 past frames for the accumulated KPConv comes from the recommendation inside the aforementioned repository.
    \item Cylinder3D: \href{https://github.com/xinge008/Cylinder3D}{xinge008/Cylinder3D}. Model: cylinder asym; learning rate: 1e-3; number of epochs: 40; batch size: 2; grid size: 480x360x32. Same parameters for nuScenes and SemanticKITTI.
    \item SRUNet and SPVCNN: \href{https://github.com/mit-han-lab/spvnas}{mit-han-lab/spvnas}. Model: MinkUnet and SPVCNN; learning rate: 2.4e-1; number of epochs: 15 for SemanticKITTI, 20 for nuScenes; batch size: 2 for SemanticKITTI, 8 for nuScenes; voxel size: 0.05m.
    \item CENet: \href{https://github.com/huixiancheng/CENet}{huixiancheng/CENet} . Model: senet-512; learning rate: 1e-2; number of epochs: 100; batch size: 6; Image resolution: 512x64. Only trained on SemanticKITTI.
    \item Helix4D: \href{https://github.com/romainloiseau/Helix4D}{romainloiseau/Helix4D}. Model: Helix4D; learning rate: 2e-3; number of epochs: 50; batch size: 2. Only trained on SemanticKITTI.
\end{itemize}

\subsection{3DLabelProp}

As mentioned in the main paper, the KPConv model used inside 3DLabelProp is the standard layout, namely the KPFCNN. It consists of 4 downsampling blocks and 4 upsampling blocks. 

In every case, the model architecture does not change. For the training, we use a Lovasz loss and weighted cross-entropy loss. We use a SGD optimizer with a weight decay of 1e-4 and a momentum of 0.98. The learning rate scheduler is a cosine annealing scheduler. It is trained with cluster extracted by the 3DLabelProp pipeline when assuming that past inferences are the ground truth. Training hyperparameters are:
\begin{itemize}
    \item SemanticKITTI - learning rate: 0.005, batch size: 12, number of iterations: 400 000.
    \item nuScenes - learning rate: 0.001, batch size: 16, number of iterations: 350 000.
\end{itemize}

Due to the annotation process of nuScenes, only 2 frame per seconds are annotated, contrary to other dataset where 10 frames per seconds are annotated. As we are only using the annotated frames for the training of the nuScenes model, we reproduce the same accumulation pattern at inference time.

In practice, we accumulate only one out of 5 frames when processing other datasets to emulate a 2Hz frame rate. It means that to predict semantic of points in frame $n$, we use accumulation of frames $n$, $n-5$, $n-10$; $n-15$; $n-20$. 

\subsection{Data augmentation}

All data augmentations applied are the same for the various methods. They are also the one used for 3DLabelProp, but it is important to note that in the 3DLabelProp case, these augmentations are applied at the cluster level and not at the scan level.

Here is the list: 
\begin{itemize}
    \item Centering.
    \item Random rotation around z.
    \item Random scaling.
    \item Gaussian noise.
    \item Flipping around x and around y.
\end{itemize}

\section{Label mappings}

The choice of labels for the different mIoU computations can have a significant impact on the results. We share in \autoref{tab:label_sk_sp} to \autoref{tab:label_ns_ps} all the details of the mappings between datasets so that all the results can be easily reproduced and that in the future new generalization methods can be compared with our results.

Some mapping can look unnecessarily coarse. For instance, SemanticKITTI define trucks as "Trucks, vans with a body that is separate from the driver cabin, pickup trucks, as well as their attached trailers.", whereas nuScenes define trucks as "Vehicles primarily designed to haul cargo including pick-ups, lorrys, trucks and semi-tractors.", which means that the sub element "van" can belong either to truck or not depending on the dataset. In order to not be sensitive to annotation choices, we have to group car and truck together into a single coarse category, otherwise we could be measuring annotation discrepancies.

This need for coarseness is exacerbated for $\mathcal{L}_{NS\cap SP}$ which is extremely coarse, and as a result can be considered an easy label set.

\begin{table}[h]
    \centering
    \begin{tabular}{c|c|c}
        SemanticKITTI & $\mathcal{L}_{SK\cap NS}$ & nuScenes\\\hline\hline

        \specialcell{motorcycle \\ motorcyclist} & motorcycle & \specialcell{motorcycle}\\\hline
        \specialcell{bicycle \\ bicyclist} & bicycle & \specialcell{bicycle}\\\hline
        \specialcell{person} & person & \specialcell{pedestrian}\\\hline
        \specialcell{road\\parking} & d. ground & \specialcell{driveable surface}\\\hline
        \specialcell{sidewalk} & sidewalk & \specialcell{sidewalk}\\\hline
        \specialcell{other ground} & o. ground & \specialcell{other flat}\\\hline
        \specialcell{building\\fence\\pole\\traffic sign} & manmade & \specialcell{barrier\\traffic cone\\manmade}\\\hline
        \specialcell{vegetation\\trunk} & vegetation & \specialcell{vegetation}\\\hline
        \specialcell{car\\truck\\other vehicle} & vehicle & \specialcell{bus\\car\\construction vehicle\\trailer\\truck}\\\hline
        \specialcell{terrain} & terrain & \specialcell{terrain}\\
    \end{tabular}
    \caption{Details of the mapping from the original label sets of SemanticKITTI and nuScenes to $\mathcal{L}_{SK\cap NS}$.}
    \label{tab:label_sk_sp}
\end{table}

\begin{table}[!h]
    \centering
    \begin{tabular}{c|c|c}
         SemanticKITTI & $\mathcal{L}_{SK\cap SP}$  & SemanticPOSS\\\hline\hline
         \specialcell{person}&person&\specialcell{person}\\ \hline
         \specialcell{bicyclist\\motorcyclist}&rider&\specialcell{rider}\\ \hline
         \specialcell{bicycle\\motorcycle}&bike&\specialcell{bike}\\ \hline
         \specialcell{other vehicle\\car\\truck}&car&\specialcell{car}\\ \hline
         \specialcell{road\\terrain\\parking\\sidewalk\\other ground}&ground&\specialcell{ground}\\ \hline
         \specialcell{trunk}&trunk&\specialcell{trunk}\\ \hline
         \specialcell{vegetation}&vegetation&\specialcell{plants}\\ \hline
         \specialcell{traffic sign}&traffic sign&\specialcell{traffic sign}\\ \hline
         \specialcell{pole}&pole&\specialcell{pole}\\ \hline
         \specialcell{building}&building&\specialcell{building\\garbage can\\cone/stone}\\ \hline
         \specialcell{fence}&fence&\specialcell{fence}\\ 

    \end{tabular}
    \caption{Details of the mapping from the original label sets of SemanticKITTI and SemanticPOSS to $\mathcal{L}_{SK\cap SP}$.}
    \label{tab:label_sk_ns}
\end{table}

\begin{table}[!h]
    \centering
    \begin{tabular}{c|c|c}
         nuScenes & $\mathcal{L}_{NS\cap SP}$  & SemanticPOSS \\ \hline\hline
         \specialcell{pedestrian}&person&\specialcell{person}\\ \hline
         \specialcell{bicycle\\motorcycle}&bike&\specialcell{rider \\ bike}\\ \hline
         \specialcell{car\\bus\\construction vehicle\\trailer\\truck}&car&\specialcell{car}\\ \hline
         \specialcell{driveable surface\\other flat\\sidewalk\\terrain}&ground&\specialcell{ground}\\ \hline
         \specialcell{vegetation}&vegetation&\specialcell{vegetation \\ plant}\\ \hline
         \specialcell{barrier\\manmade\\traffic cone}&manmade&\specialcell{traffic sign\\pole\\garbage can\\building\\cone/stone\\fence}\\ 

    \end{tabular}
    \caption{Details of the mapping from the original label sets of nuScenes and SemanticPOSS to $\mathcal{L}_{NS\cap SP}$.}
    \label{tab:label_ns_sp}
\end{table}

\begin{table}[!h]
    \centering
    \begin{tabular}{c|c|c}
        SemanticKITTI & $\mathcal{L}_{SK\cap PS}$ & PandaSet\\\hline\hline
         \specialcell{bicycle\\motorcycle\\bicyclist\\motorcyclist}&2-wheeled&\specialcell{pedicab\\personal mobility device\\motorized scooter\\bicycle\\motorcycle}\\ \hline
         \specialcell{person}&pedestrian&\specialcell{pedestrian\\pedestrian w/ objects}\\ \hline
         \specialcell{road\\parking}&d. ground&\specialcell{road\\road marking\\driveway}\\ \hline
         \specialcell{sidewalk}&sidewalk&\specialcell{sidewalk}\\ \hline
         \specialcell{other ground\\terrain}&o. ground&\specialcell{ground}\\ \hline
         \specialcell{building\\fence\\pole\\traffic sign}&manmade&\specialcell{pylons\\road barriers\\signs\\cones\\construction signs\\construction barriers\\rolling containers\\building\\other static object}\\ \hline
         \specialcell{vegetation\\trunk}&vegetation&\specialcell{vegetation}\\ \hline
         \specialcell{car\\truck\\other vehicle}&4-wheeled&\specialcell{pickup truck\\medium sized\\truck\\semi-truck\\towed object\\construction vehicle\\uncommon vehicle\\emergency vehicle\\bus\\car}\\ 
    \end{tabular}
    \caption{Details of the mapping from the original label sets of SemanticKITTI and PandaSet to $\mathcal{L}_{SK\cap PS}$.}
    \label{tab:label_sk_ps}
\end{table}
\clearpage
\begin{table}[!h]
    \centering
    \begin{tabular}{c|c|c}
         nuScenes & $\mathcal{L}_{NS\cap PS}$  & PandaSet \\ \hline\hline
         \specialcell{bicycle\\ motorcycle}&2-wheeled&\specialcell{pedicab\\personal mobility device\\motorized scooter\\bicycle\\motorcycle}\\ \hline
         \specialcell{pedestrian}&pedestrian&\specialcell{pedestrian\\pedestrian w/ objects}\\ \hline
         \specialcell{driveable ground}&d. ground&\specialcell{road\\road marking\\driveway}\\ \hline
         \specialcell{sidewlak}&sidewalk&\specialcell{sidewalk}\\ \hline
         \specialcell{other flat\\terrain}&o. ground&\specialcell{ground}\\ \hline
         \specialcell{barrier\\manmade\\traffic cone}&manmade&\specialcell{pylons\\road barriers\\signs\\cones\\construction signs\\construction barriers\\rolling containers\\building\\other static object}\\ \hline
         \specialcell{vegetation}&vegetation&\specialcell{vegetation}\\ \hline
         \specialcell{car\\bus\\construction vehicle\\trailer\\truck}&4-wheeled&\specialcell{pickup truck\\medium sized\\truck\\semi-truck\\towed object\\construction vehicle\\uncommon vehicle\\emergency vehicle\\bus\\car}\\ 
    \end{tabular}
    \caption{Details of the mapping from the original label sets of nuScenes and PandaSet to $\mathcal{L}_{NS\cap PS}$.}
    \label{tab:label_ns_ps}
\end{table}
\newpage

\section{Quantitative \& Qualitative results}

 In \autoref{tab:sk} to \autoref{tab:ns_ns_sp}, we present the details by class of all semantic segmentation results presented in the main article. They are complementary of \autoref{tab:DG_from_SK} and \autoref{tab:DG_from_NS}

As mentioned in the main article, \autoref{tab:DG_SK_NS} we see the ability of KPConv to perform well on the Person class (+12\% over the second best), which explains KPConv outranking every other methods. We could believe that 3DLabelProp has trouble reproducing such a high quality of dynamic objects due to the accumulation process, which leaves trail inside the scan. Nonetheless, we did try a variant of 3DLabelProp by discarding past dynamic object (to remove trails), and we notice a decrease in general performance (especially "mobile" objects like parked cars).

We see the impact of the coarseness of $\mathcal{L}_{NS\cap SP}$ in \autoref{tab:ns_ns_sp} as all methods are well performing.

In the various result tables, we did not include the number of parameters of the various models, even if it is an important variable in real time semantic segmentation. This choice was made because, from one segmentation paradigm to another, the parameters don't have the same impact.

\begin{figure*}
        \centering
        \includegraphics[width=\linewidth]{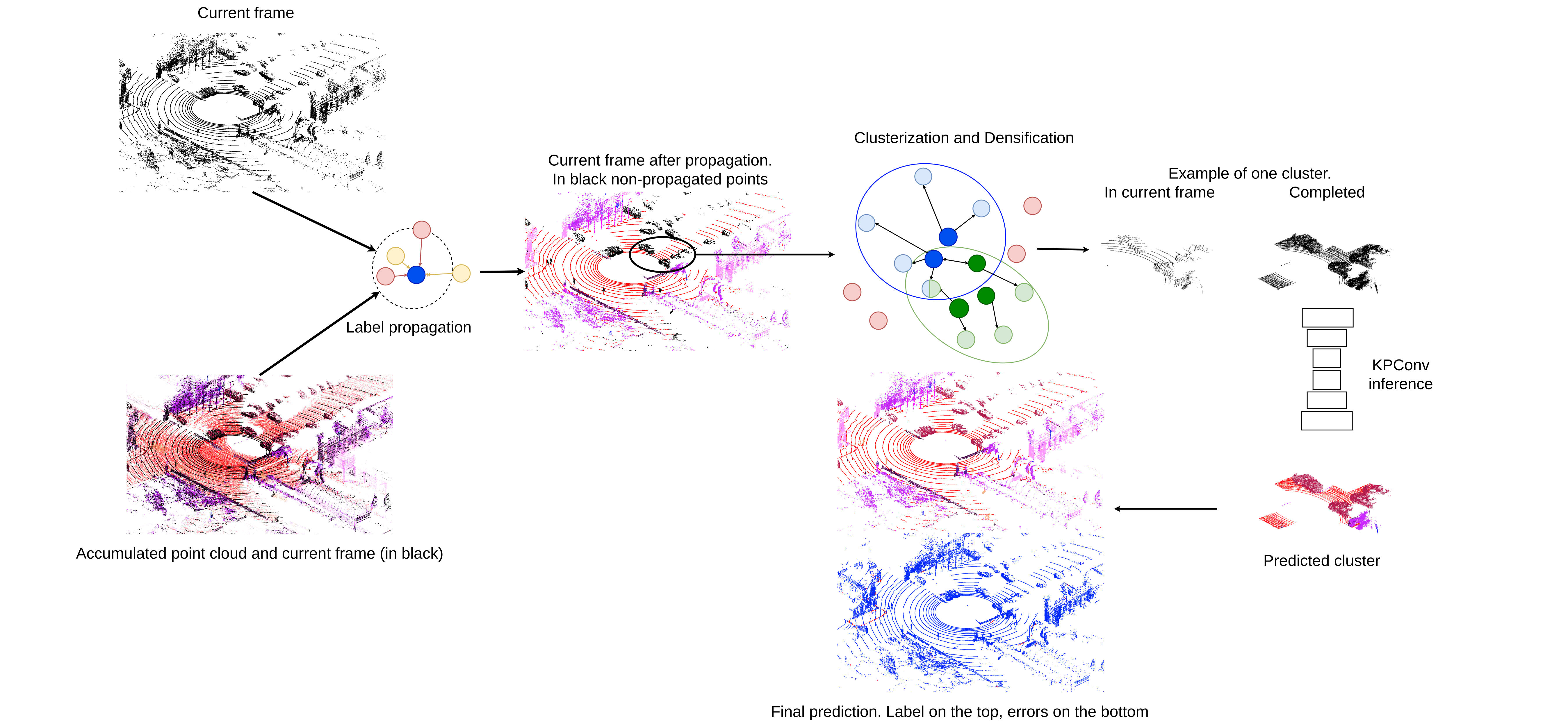}
        \caption{Example of the 3DLabelProp pipeline. Trained on SemanticKITTI and tested on SemanticPOSS.}
        \label{fig:3dlab}
\end{figure*}

At the end of this part in \autoref{fig:3dlab}, we illustrate the 3DlabelProp process.

\begin{table*}[!h]
\tiny
    \centering
    \begin{tabular}{c|c|c|c|c|c|c|c|c|c|c|c|c|c|c|c|c|c|c|c||c}
          & \rotatebox{90}{Car} & \rotatebox{90}{Bicycle} & \rotatebox{90}{Motorcycle}& \rotatebox{90}{Truck}&\rotatebox{90}{Other-vehicle} &\rotatebox{90}{Person} &\rotatebox{90}{Bicyclist} &\rotatebox{90}{Motrocyclist} & \rotatebox{90}{Road}& \rotatebox{90}{Parking}&\rotatebox{90}{sidewalk} & \rotatebox{90}{Other-ground}&\rotatebox{90}{building} & \rotatebox{90}{Fence}&\rotatebox{90}{Vegetation} &\rotatebox{90}{Trunk} & \rotatebox{90}{Terrain}& \rotatebox{90}{Pole}& \rotatebox{90}{Traffic-sign}& \rotatebox{90}{mIoU} \\ \hline
        CENet & 91.3& 24.8& 60.6& 81.8& 57.1& 56.9& 76.5&0.0&93.2& 51.6& 78.7& 0.1& 84.6& 54.9& 82.9& 60.1& 68.6& 54.6& 38.1& 58.8 \\\hline
        Helix4D   & 96.0& 21.2& 63.5& 68.2& 61.4& 64.9& 73.9& 0.5& 93.3& 39.4& 79.2& 1.4& 88.6& 54.1& 88.5& 60.7& 77.4& 61.2& 46.0& 60.0 \\\hline
        KPConv  & 94.7 & 36.8& 58.6& 45.1& 47.7& 62.7& 76.8& 1.0& 90.1& 32.3& 75.5& 3.8& 88.4& 59.8& 87.6& 67.2& 74.3& 61.7& 44.4& 58.3   \\\hline
        SRUNet   & 96.2 & 7.9 & 51.6& 65.5& 51.5& 64.7& 64.5&0.0&93.2& 48.3& 80.1& 0.1& 91.0& 62.3& 88.4& 67.7& 75.4& 62.8& 43.3& 58.6 \\\hline
        SPVCNN   & 96.5& 22.6& 59.4& 79.4& 61.8& 67.1& 81.0&0.0&93.3& 47.7& 80.5& 0.2& 91.1& 63.8& 87.7& 66.9& 73.2& 63.5& 47.7&62.3  \\\hline
        Cylinder3D   & 96.4 & 21.0& 53.2& 79.1& 57.4& 69.2& 80.5&0.0&93.2& 43.8& 79.4& 2.6& 90.4& 55.7& 86.0& 64.4&71.6& 63.5&45.8 &60.7  \\\hline 
\hline
        3DLabelProp (Ours) & 96.4& 38.7& 80.3& 63.5& 62.6& 76.5& 92.6&0.0&85.9& 45.6&66.5 &4.2& 87.4& 49.0& 86.4& 65.6&70.0&48.5&36.4&60.8\\
    \end{tabular}
    \caption{Source-to-source segmentation results for SemanticKITTI on $\mathcal{L}_{SK}$.}
    \label{tab:sk}
\end{table*}

\begin{table*}[!h]
\tiny
    \centering
    \begin{tabular}{c|c|c|c|c|c|c|c|c|c|c|c|c|c|c|c|c|c|c|c||c}
          & \rotatebox{90}{Car} & \rotatebox{90}{Bicycle} & \rotatebox{90}{Motorcycle}& \rotatebox{90}{Truck}&\rotatebox{90}{Other-vehicle} &\rotatebox{90}{Person} &\rotatebox{90}{Bicyclist} &\rotatebox{90}{Motrocyclist} & \rotatebox{90}{Road}& \rotatebox{90}{Parki-ng}&\rotatebox{90}{sidewalk} & \rotatebox{90}{Other-ground}&\rotatebox{90}{building} & \rotatebox{90}{Fence}&\rotatebox{90}{Vegetation} &\rotatebox{90}{Trunk} & \rotatebox{90}{Terrain}& \rotatebox{90}{Pole}& \rotatebox{90}{Traffic-sign}& \rotatebox{90}{mIoU} \\ \hline
        CENet & 82.7& 1.8& 20.8& 38.2& 16.4& 13.1& 51.2&0.0&86.7& 18.5& 67.4& 0.2& 69.7& 32.6& 76.3& 38.4& 69.9& 46.5& 11.7 & 39.1 \\\hline
        Helix4D   & 93.6& 20.3& 42.1& 35.0& 52.7& 56.1& 63.4& 2.7& 89.6& 24.9& 73.4& 3.3& 86.3& 48.4& 86.8& 56.3& 73.7& 58.2& 43.5& 53.2 \\\hline
        KPConv  & 92.4 & 30.6& 52.6& 27.4& 30.1& 55.5& 70.5& 2.0& 84.4& 21.6& 67.2& 8.4& 83.7& 56.9& 87.3& 60.1& 74.1& 58.6& 37.8& 52.7   \\\hline
        SRUNet   & 94.0 & 7.3& 45.1& 57.0& 43.9& 54.8& 53.6&0.0&90.7& 37.1& 75.5& 1.4& 88.9& 54.9& 87.6& 62.8& 74.1& 60.1& 37.2& 54.0 \\\hline
        SPVCNN   & 94.0& 21.4& 46.3& 72.7& 52.7& 60.4& 67.6&0.0&90.7& 37.3& 75.4& 1.9& 89.1& 57.9& 87.3& 62.8& 72.5& 59.2& 41.2& 57.4 \\\hline
        Cylinder3D   & 92.7 &20.8 &38.4 &65.2 &48.7 &56.5 &64.5& 0.0&90.4 &18.6 &73.1 &1.4 &88.0 &40.6 &83.4 &58.1 & 70.2& 59.3& 38.8& 53.1 \\\hline 
\hline
        3DLabelProp (Ours) & 96.2 & 35.4& 86.4& 63.8& 66.7& 71.7& 87.8& 0.0& 77.8& 38.7& 59.3& 0.7& 83.5& 42.6& 82.6& 53.6& 56.3& 43.3& 31.0& 56.7 \\
    \end{tabular}
    \caption{Validation set results on generalization from SemanticKITTI to SemanticKITTI32 on $\mathcal{L}_{SK}$.}
    \label{tab:sk_sk32}
\end{table*}

\begin{table*}[ht]
\scriptsize
    \centering
    \begin{tabular}{c|c|c|c|c|c|c|c|c||c}
 & 2-wheeled & Pedestrian & D. Ground & Seidewalk & O.Ground & Manmade & Vegetation & 4-Wheeled & mIoU    \\\hline
        CENet &0.1&0.7&5.2&7.2&7.0&51.0&28.2&7.4&13.3  \\\hline
        Helix4D &07.6&13.9&41.7&23.7&6.4&54.6&32.2&42.5&27.7\\\hline
        KPConv &6.4&24.6&44.9&25.7&15.2&51.3&40.5&53.0&32.7\\\hline
        SRUNet &17.8&34.7&54.8&31.8&16.9&77.7&57.9&61.6&44.2 \\\hline
        SPVCNN &11.0&23.8&51.2&33.9&17.5&77.1&53.4&53.4&40.2\\\hline
        Cylinder3D &2.7&11.9&15.5&18.4&10.4&52.2&25.1&10.9&18.4\\\hline\hline
        3DLabelProp (Ours) &33.0&69.7&52.3&37.3&16.3&83.5&68.2&87.3&56.0  \\
    \end{tabular}
    \caption{Validation set results on generalization from SemanticKITTI to Panda64 on the label set $\mathcal{L}_{SK\cap PS}$.}
    \label{tab:DG_P64}
\end{table*}

\begin{table*}[ht]
\scriptsize
    \centering
    \begin{tabular}{c|c|c|c|c|c|c|c|c||c}
 & 2-wheeled & Pedestrian & D. Ground & Seidewalk & O.Ground & Manmade & Vegetation & 4-Wheeled & mIoU    \\\hline
        CENet &0.0&0.0&4.9&1.4&2.0&11.1&21.4&3.1&4.9  \\\hline
        Helix4D &3.5&4.0&24.0&5.4&4.7&28.4&20.4&23.5&14.2\\\hline
        KPConv &10.1&14.6&13.7&7.0&13.1&20.1&42.2&48.0&21.1\\\hline
        SRUNet &5.6&12.1&29.4&7.2&7.1&41.9&42.2&32.2&22.2  \\\hline
        SPVCNN &2.0&6.7&28.7&7.6&6.1&49.9&37.5&16.9&19.4\\\hline
        Cylinder3D &0.0&1.4&1.9&4.8&8.4&20.9&13.2&1.4&6.5\\\hline\hline
        3DLabelProp (Ours) &41.9&61.8&65.2&27.0&10.1&73.9&72.6&89.5&55.2  \\
    \end{tabular}
    \caption{Validation set results on generalization from SemanticKITTI to PandaFF on the label set $\mathcal{L}_{SK\cap PS}$.}
    \label{tab:DG_PFF}
\end{table*}

\begin{table*}[ht]
\scriptsize
    \centering
    \begin{tabular}{c|c|c|c|c|c|c|c|c|c|c|c||c}
         & Person & Rider & Bike & Car & Ground & Trunk & Vegetation & Traffic-sign & Pole & Building & Fence & mIoU  \\\hline

        CENet &2.9&40.2&0.0&31.9&73.9&33.3&65.6&22.2&36.5&77.6&37.0&27.9  \\\hline
        Helix4D &27.8&18.4&4.0&59.7&64.4&27.7&58.0&24.3&29.2&66.5&16.4&36.0\\\hline
        KPConv &43.9&32.7&8.6&63.4&74.8&31.4&59.7&7.6&32.6&57.6&17.6&39.1\\\hline
        SRUNet &38.9&26.1&2.9&82.9&75.1&35.5&65.6&15.5&41.6&77.1&37.3&45.3  \\\hline
        SPVCNN &44.6&18.1&4.6&83.2&75.0&36.0&65.8&18.5&41.6&75.8&36.0&45.4\\\hline
        Cylinder3D &44.6&13.6&2.8&81.5&73.6&33.2&61.6&13.1&36.0&71.3&30.1&41.9\\\hline\hline
        3DLabelProp (Ours) &71.6&43.9&3.6&89.3&73.9&33.3&65.6&22.2&36.5&77.6&37.0&50.4  \\

    \end{tabular}
    \caption{Validation set results on generalization from SemanticKITTI to SemanticPOSS on the label set $\mathcal{L}_{SK\cap SP}$. }
    \label{tab:DG_SP}
\end{table*}

\begin{table*}[ht]
\scriptsize
    \centering
    \begin{tabular}{c|c|c|c|c|c|c|c|c|c|c||c}
 & Motorcycle & Bicycle & Person & Road & Sidewalk & O. Ground & Manmade & Vegetation & Car & Terrain & mIoU    \\\hline
        CENet &0.0&0.0&0.0&0.8&0.2&1.3&24.6&16.8&6.0&0.2&5.0\\\hline
        Helix4D &7.5&2.8&23.2&75.7&31.7&3.8&57.2&58.5&53.7&28.5&34.3\\\hline
        KPConv &33.7&7.8&52.8&80.3&34.9&4.0&70.7&70.5&74.3&37.8&46.7\\\hline
        SRUNet &17.1&2.4&24.8&87.8&39.4&5.1&74.7&72.6&78.5&25.5&42.8\\\hline
        SPVCNN &35.9&3.0&27.3&88.3&39.6&6.6&75.0&72.4&78.2&24.7&45.1\\\hline
        Cylinder3D &3.3&2.2&1.0&77.3&32.0&7.6&67.3&60.6&56.2&19.8&32.7\\\hline\hline
        3DLabelProp (Ours) &27.4&8.8&40.9&79.3&38.3&5.0&69.8&69.4&72.8&32.7&44.4 \\
    \end{tabular}
    \caption{Validation set results on generalization from SemanticKITTI to nuScenes on the label set $\mathcal{L}_{SK\cap NS}$.}
    \label{tab:DG_SK_NS}
\end{table*}

\begin{table*}[!h]
\scriptsize
    \centering
    \begin{tabular}{c|c|c|c|c|c|c|c|c|c|c|c|c|c|c|c|c||c}
        & \rotatebox{90}{barrier}& \rotatebox{90}{bicycle}& \rotatebox{90}{bus}& \rotatebox{90}{car}& \rotatebox{90}{cnstrctn-vhcl}& \rotatebox{90}{motorcycle}& \rotatebox{90}{pedestrian}& \rotatebox{90}{traffic-cone}& \rotatebox{90}{trailer}& \rotatebox{90}{truck}& \rotatebox{90}{drvbl-grnd}& \rotatebox{90}{other-flat}& \rotatebox{90}{sidewalk}&\rotatebox{90}{terrain} & \rotatebox{90}{manmade}& \rotatebox{90}{vegetation}& \rotatebox{90}{mIoU} \\ \hline
        KPConv &66.2 &17.8 &72.4 &87.8 &26.3 &67.8 &69.8 &51.5 &25.6 &73.5 &93.8 &53.6 &66.3 &71.2 &83.0 &82.8 &63.1  \\ \hline
        SPVCNN & 72.6& 14.0& 82.9& 88.7& 32.2& 73.4& 69.9& 47.8& 46.1& 78.3& 95.0& 64.4& 69.4& 71.8& 84.6& 83.4& 67.2  \\\hline
        Cylinder3D& 71.5 &29.4 &84.3 &86.4 &40.5 &70.5 &72.9 &54.3 &57.2 &79.7 &96.1 &65.8 &71.9 &71.6 &86.4 &85.0 &70.2 \\  \hline\hline
        3DLabelProp (Ours) & 71.8 & 46.9& 84.1& 84.5& 39.7& 83.1& 77.0& 50.4& 47.9& 79.7& 93.1& 62.5& 68.5& 72.9& 87.7& 86.3& 71.0  \\\
    \end{tabular}
    \caption{Source-to-source segmentation results for nuScenes on $\mathcal{L}_{NS}$.}
    \label{tab:ns}
\end{table*}

\begin{table*}[ht]
\scriptsize
    \centering
    \begin{tabular}{c|c|c|c|c|c|c|c|c|c|c||c}
 & Motorcycle & Bicycle & Person & Road & Sidewalk & O. Ground & Manmade & Vegetation & Car & Terrain & mIoU    \\\hline
        KPConv &18.9&3.9&49.8&67.2&28.8&0.4&73.4&81.8&79.0&45.8&44.9\\\hline
        SPVCNN &25.7&11.6&47.7&69.8&46.3&0.0&77.1&81.0&90.0&45.2&49.4\\\hline
        Cylinder3D &3.7&0.0&21.0&54.3&21.5&0.0&65.0&70.9&46.1&34.4&31.7\\\hline\hline
        3DLabelProp (Ours) &36.8&37.2&66.1&77.9&61.1&0.4&81.6&84.3&93.2&66.7&60.5  \\
    \end{tabular}
    \caption{Validation set results on generalization from nuScenes to SemanticKITTI on the label set $\mathcal{L}_{NS\cap SK}$.}
    \label{tab:DG_NS_SK}
\end{table*}

\begin{table*}[ht]
\scriptsize
    \centering
    \begin{tabular}{c|c|c|c|c|c|c|c|c|c|c||c}
 & Motorcycle & Bicycle & Person & Road & Sidewalk & O. Ground & Manmade & Vegetation & Car & Terrain & mIoU    \\\hline
        KPConv &23.1&3.5&46.3&78.9&48.6&1.9&75.9&83.7&81.1&62.7&50.6\\\hline
        SPVCNN &28.6&8.3&45.0&81.4&50.9&0.2&81.6&84.5&91.5&60.0&53.2\\\hline
        Cylinder3D &14.0&5.6&37.5&75.8&48.0&0.0&74.5&77.9&78.8&52.0&46.4\\\hline\hline
        3DLabelProp (Ours) &45.9&39.1&63.7&83.3&63.7&0.3&81.8&85.2&93.6&68.1&62.5  \\
    \end{tabular}
    \caption{Validation set results on generalization from nuScenes to SemanticKITTI32 on the label set $\mathcal{L}_{NS\cap SK}$.}
    \label{tab:DG_NS_SK32}
\end{table*}

\begin{table*}[ht]
\scriptsize
    \centering
    \begin{tabular}{c|c|c|c|c|c|c|c|c||c}
 & 2-wheeled & Pedestrian & D. Ground & Seidewalk & O.Ground & Manmade & Vegetation & 4-Wheeled & mIoU    \\\hline
        KPConv&3.9&6.2&20.3&17.1&10.8&64.5&53.2&24.2&25.0\\\hline
        SPVCNN &19.2&43.1&38.0&30.6&15.3&71.9&62.1&69.4&43.6\\\hline
        Cylinder3D&0.3&0.8&13.3&6.9&7.8&58.9&30.5&8.3&15.8 \\\hline\hline
        3DLabelProp (Ours) &48.9&71.5&71.9&49.8&28.6&89.2&76.9&86.1&65.4  \\
    \end{tabular}
    \caption{Validation set results on generalization from nuScenes to Panda64 on the label set $\mathcal{L}_{NS\cap PS}$.}
    \label{tab:DG_P64_NS}
\end{table*}

\begin{table*}[ht]
\scriptsize
    \centering
    \begin{tabular}{c|c|c|c|c|c|c|c|c||c}
 & 2-wheeled & Pedestrian & D. Ground & Seidewalk & O.Ground & Manmade & Vegetation & 4-Wheeled & mIoU    \\\hline
        KPConv &5.0&6.8&7.0&7.5&4.6&38.2&48.7&17.8&16.9\\\hline
        SPVCNN &1.4&7.4&3.0&3.2&8.0&28.9&28.5&8.5&11.1\\\hline
        Cylinder3D &0.0&0.0&1.6&0.6&2.4&24.3&7.8&0.9&4.7\\\hline\hline
        3DLabelProp (Ours) &51.6&68.2&90.0&40.8&28.5&85.6&81.2&87.6&66.7  \\
    \end{tabular}
    \caption{Validation set results on generalization from nuScenes to PandaFF on the label set $\mathcal{L}_{NS\cap PS}$.}
    \label{tab:DG_PFF_NS}
\end{table*}

\begin{table*}[!h]
\scriptsize
    \centering
        \begin{tabular}{c|c|c|c|c|c|c||c}
         & Person & Bike & Car & Ground & Vegetation & Manmade& mIoU \\ \hline
        KPConv  &56.0&17.2&72.4  &73.8  &70.0  & 74.5& 60.7\\\hline
        SPVCNN  &64.5&15.5& 85.6 &74.2  &71.8  &78.1 & 64.8\\\hline
        Cylinder3D  &22.9&0.6& 29.5 &74.1  &63.0  &66.9 & 42.8 \\\hline\hline
        3DLabelProp (Ours)  &72.1&0.9& 79.4 & 73.3 & 70.8 & 80.9& 64.3\\
    \end{tabular}
    \caption{Validation set results on generalization from nuScenes to SemanticPOSS on the label set $\mathcal{L}_{NS\cap SP}$. }
    \label{tab:ns_ns_sp}
\end{table*}
\captionsetup[subfigure]{labelformat=empty}

\begin{figure*}
        \centering
        \subfloat{
            \includegraphics[width=.25\linewidth]{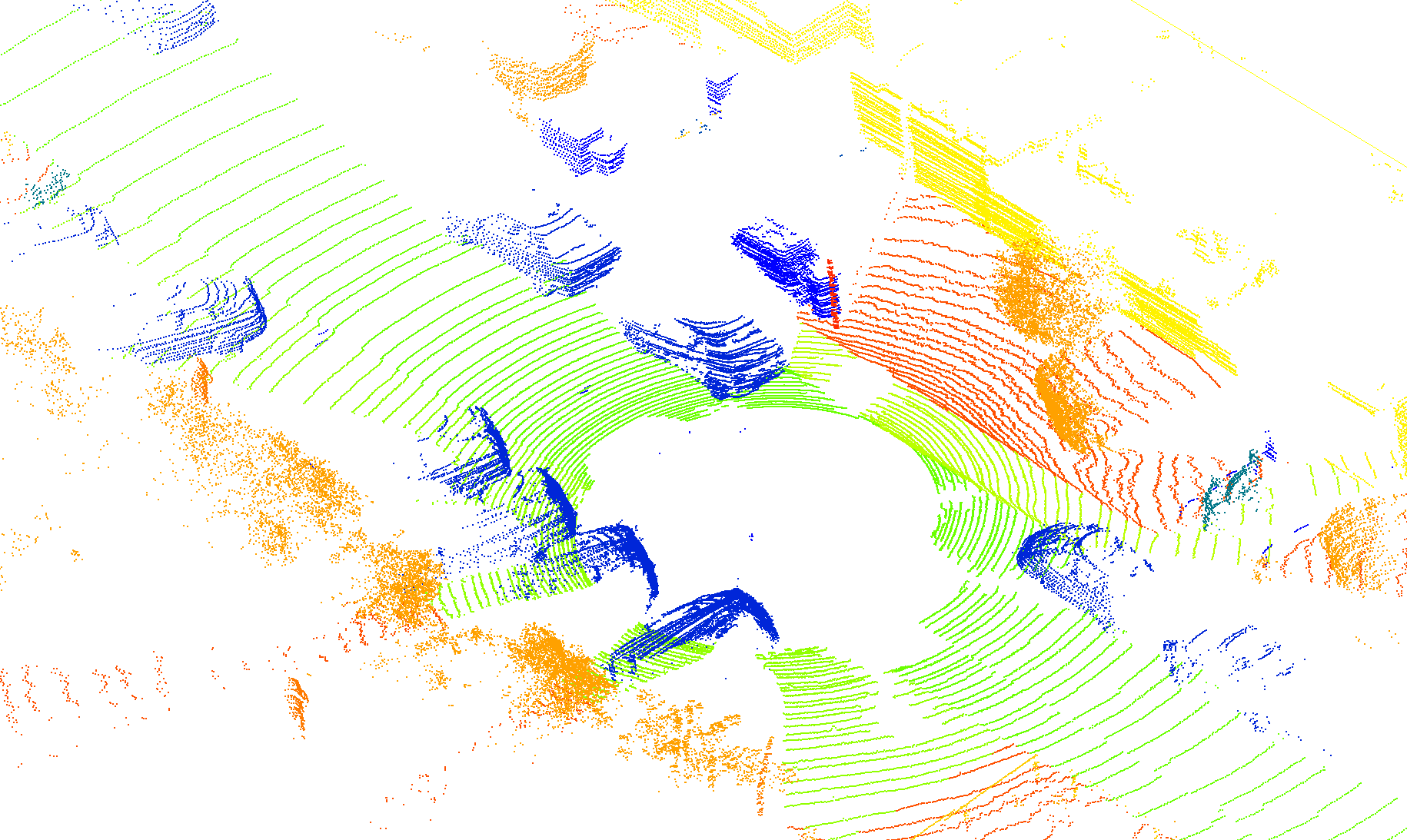}
        }
        \subfloat{
            \includegraphics[width=.25\linewidth]{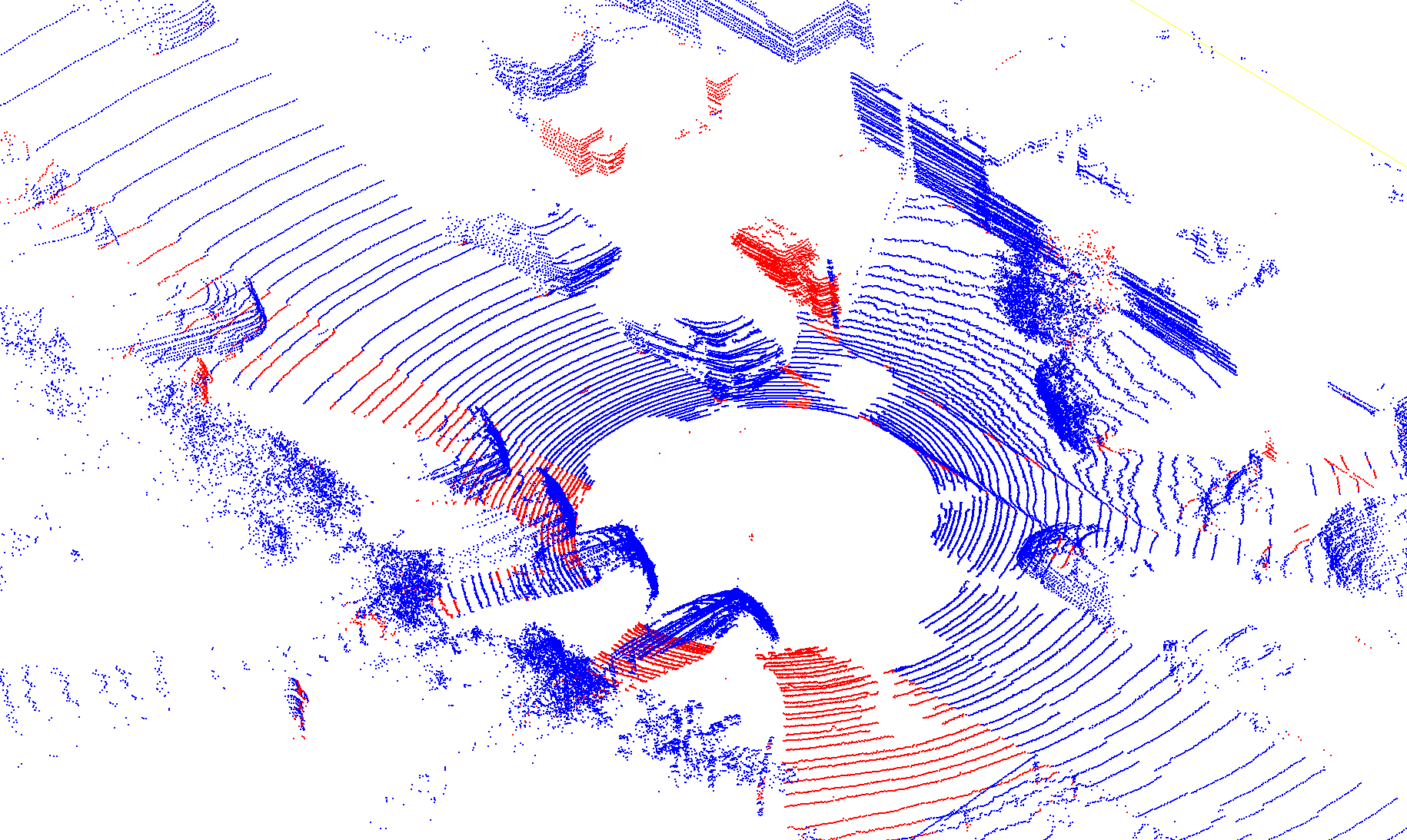}
        } 
        \subfloat{
            \includegraphics[width=.25\linewidth]{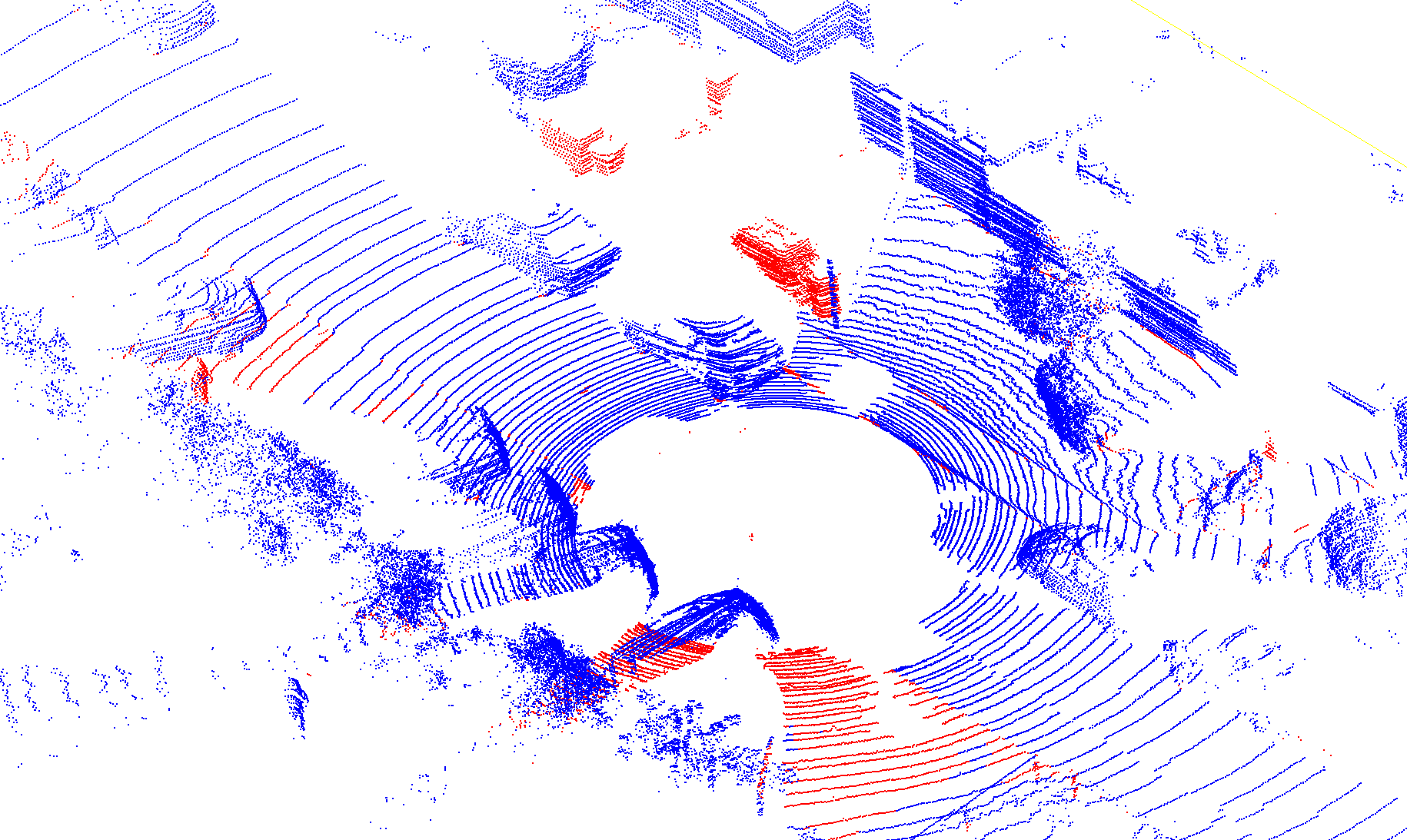}
        }
        \subfloat{
            \includegraphics[width=.25\linewidth]{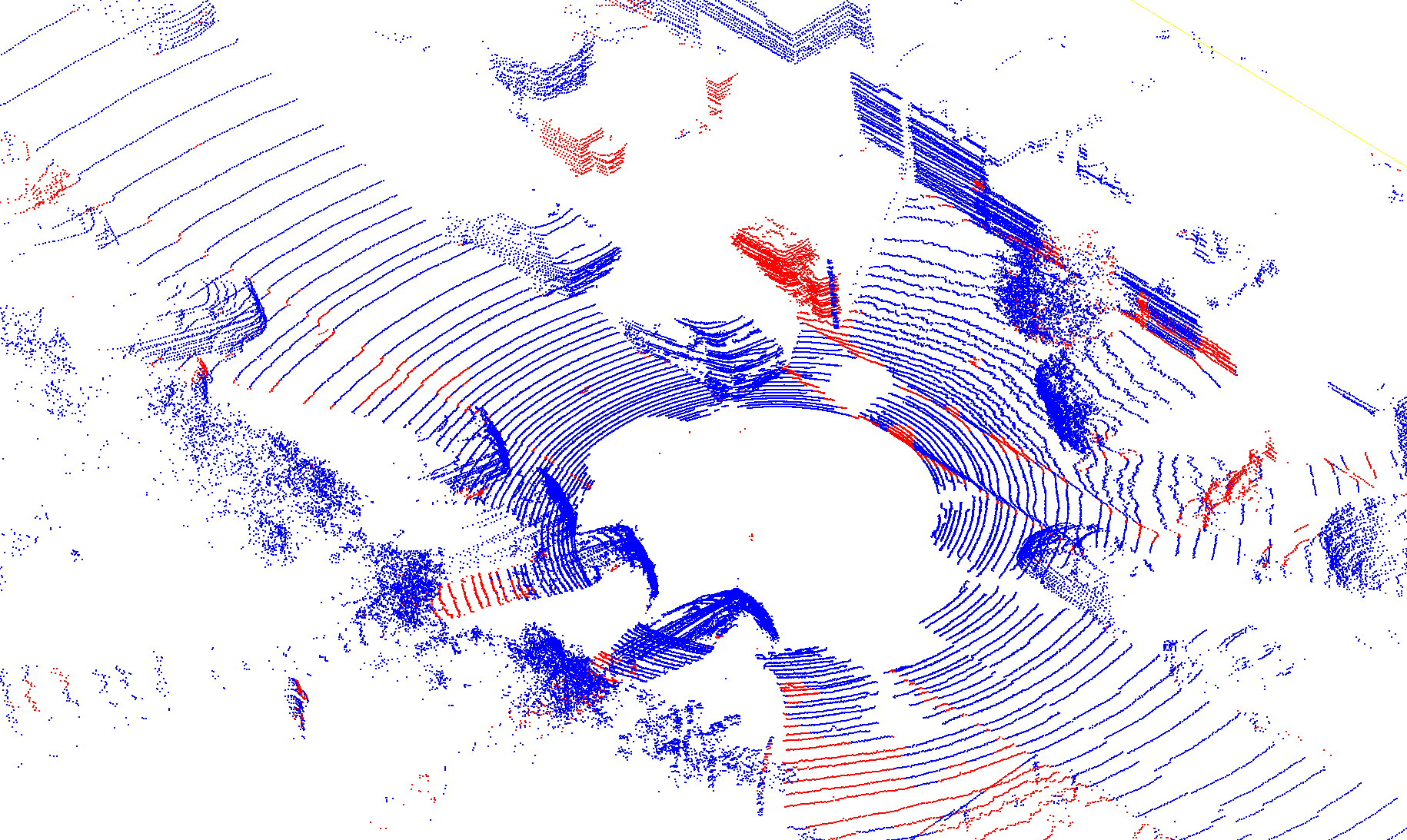}
        } 
        
        \subfloat[Ground truth]{
            \includegraphics[width=.25\linewidth]{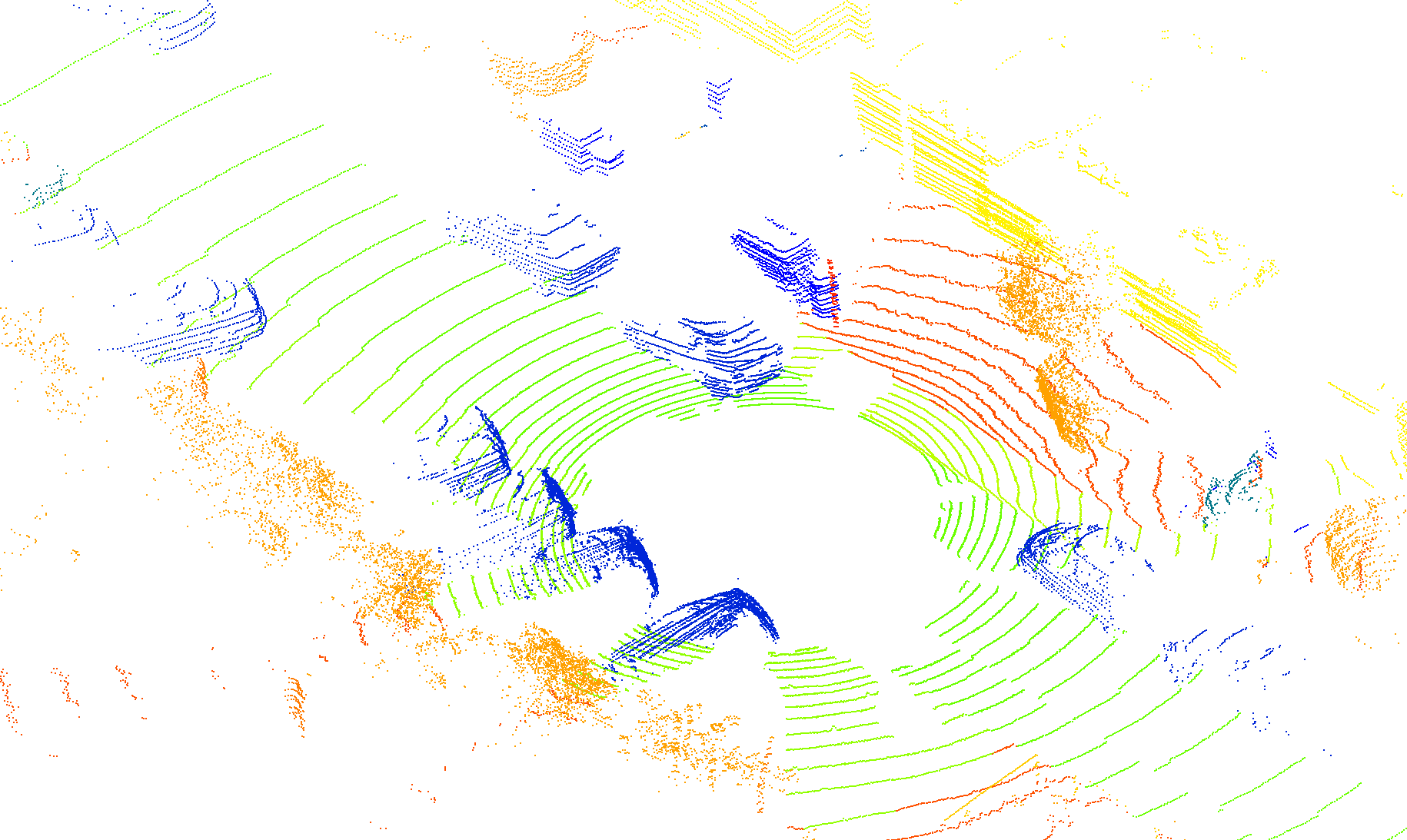}
        }
        \subfloat[KPConv]{
            \includegraphics[width=.25\linewidth]{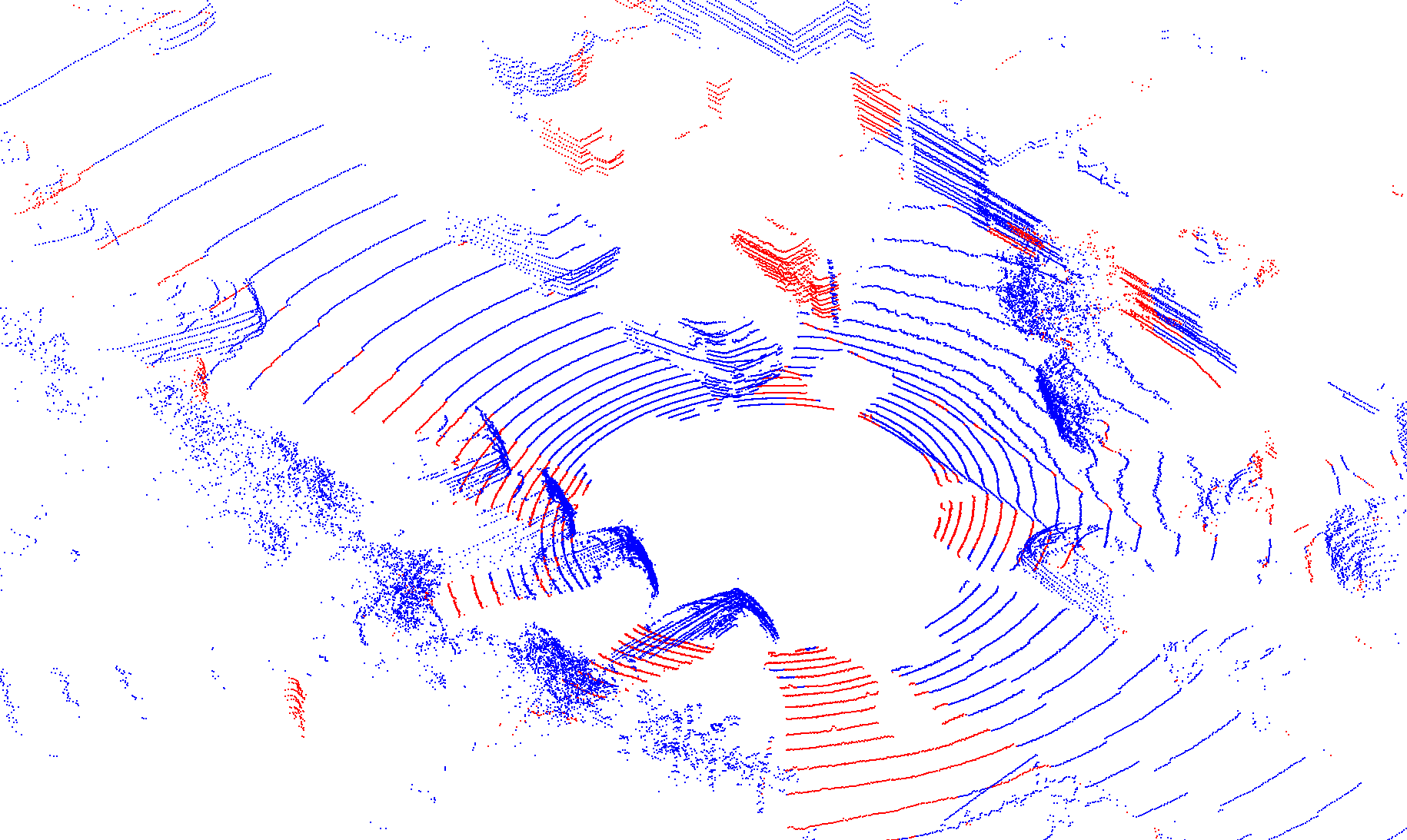}
        } 
        \subfloat[SPVCNN]{
            \includegraphics[width=.25\linewidth]{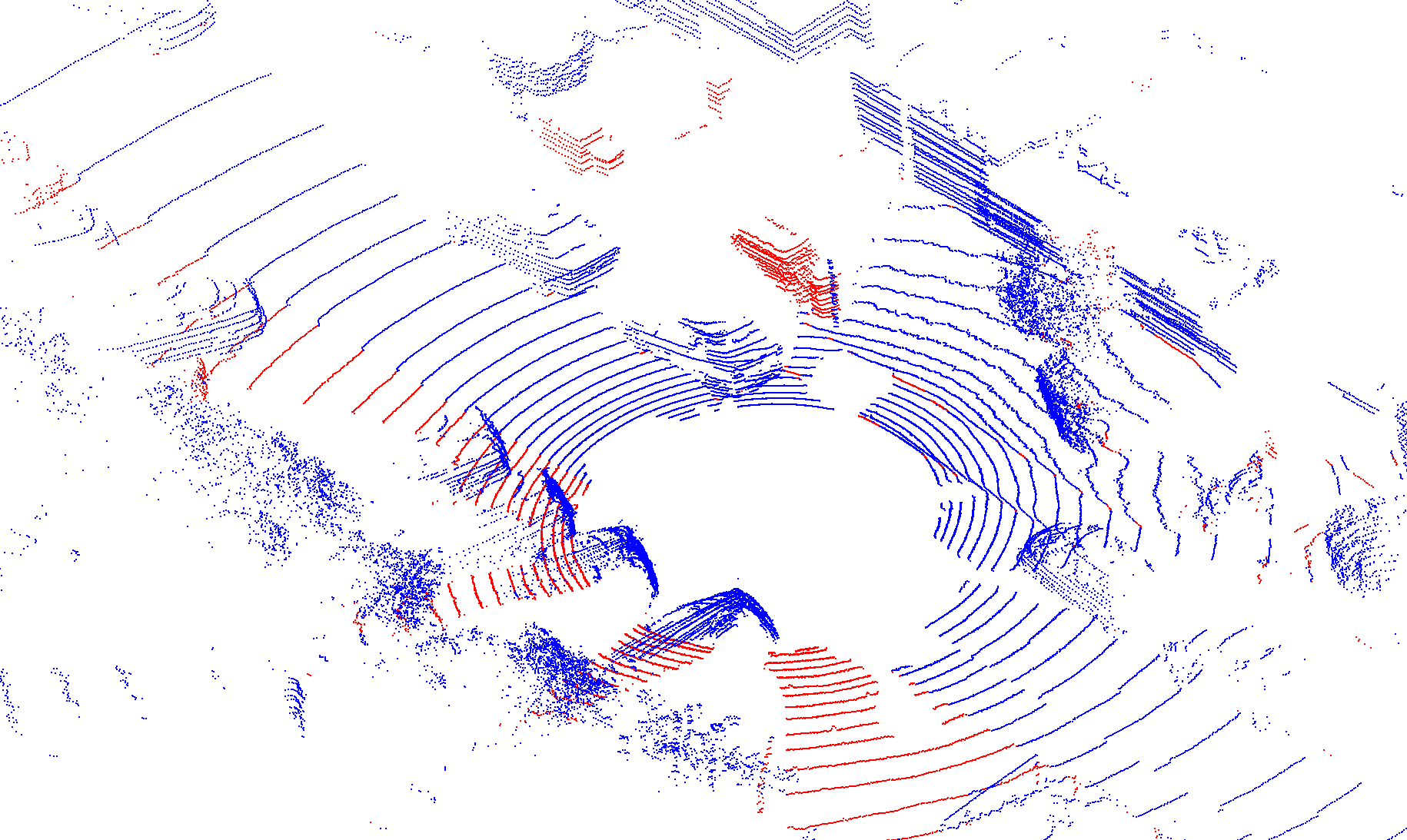}
        }
        \subfloat[3DLabelProp (Ours)]{
            \includegraphics[width=.25\linewidth]{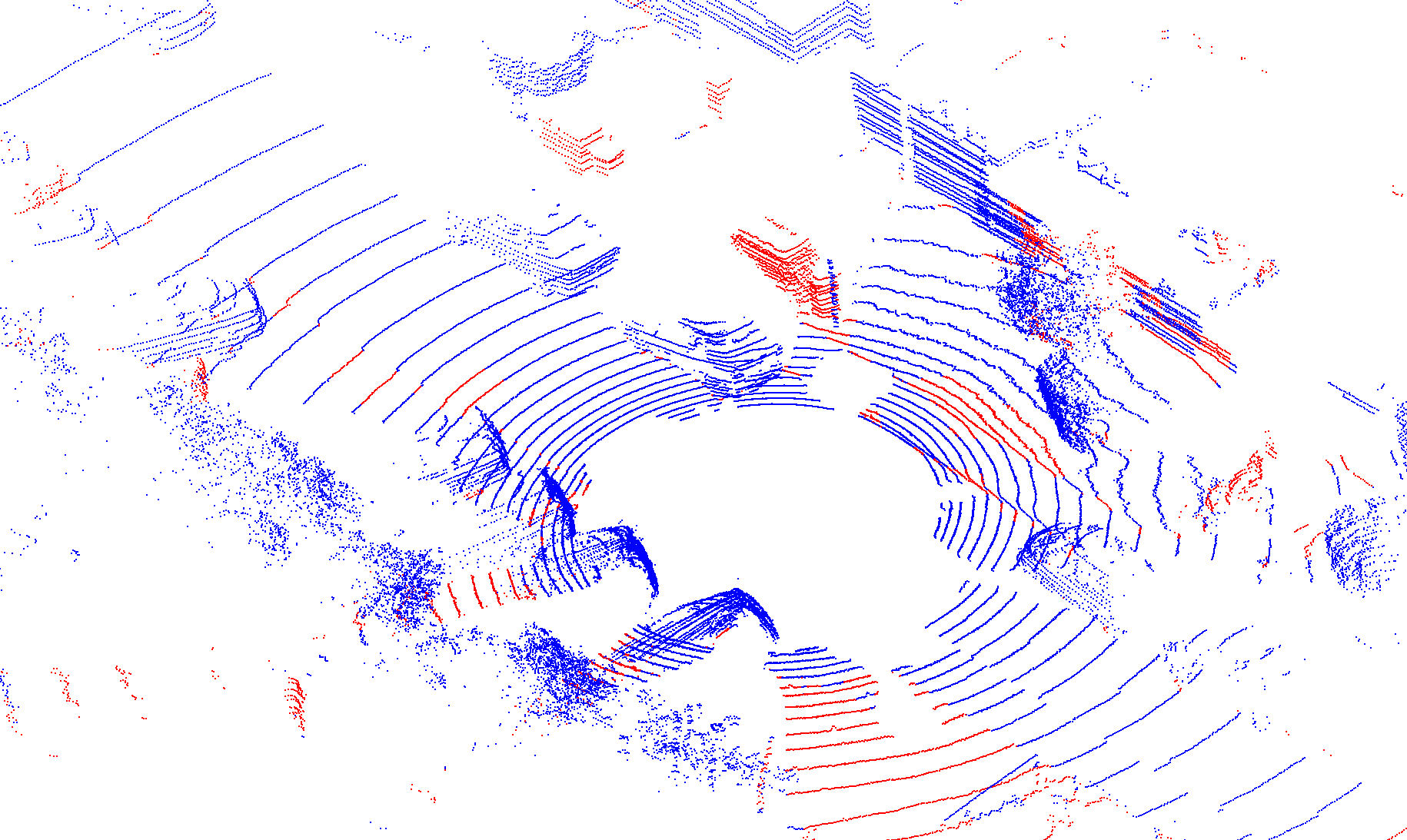}
        } 
        \caption{Qualitative results when trained on SemanticKITTI and tested on SemanticKITTI (top row) and SemanticKITTI32 (bottom row). For illustration, the same frame was selected for SemanticKITTI and SemanticKITTI32. From left to right: Ground truth labels, results from KPConv, results from SPVCNN, results from 3DLabelProp. In blue, points with correct semantic segmentations. In red, errors. \\ We can observe the decrease in performance from SemanticKITTI to SemanticKITTI32 of SPVCNN on the left sidewalk which is badly labeled for SemanticKITTI32. For KPConv, the buildings on the right are errors appearing only with SemanticKITTI32. For 3DLabelProp, while the segmentation is slightly worse than SPVCNN on the full resolution, because the decrease of performance is small between SemanticKITTI and SemanticKITTI32, the overall quality for SemanticKITTI32 is on par with SPVCNN.}
\end{figure*}

\begin{figure*}
        \centering
        \subfloat{
            \includegraphics[width=.25\linewidth]{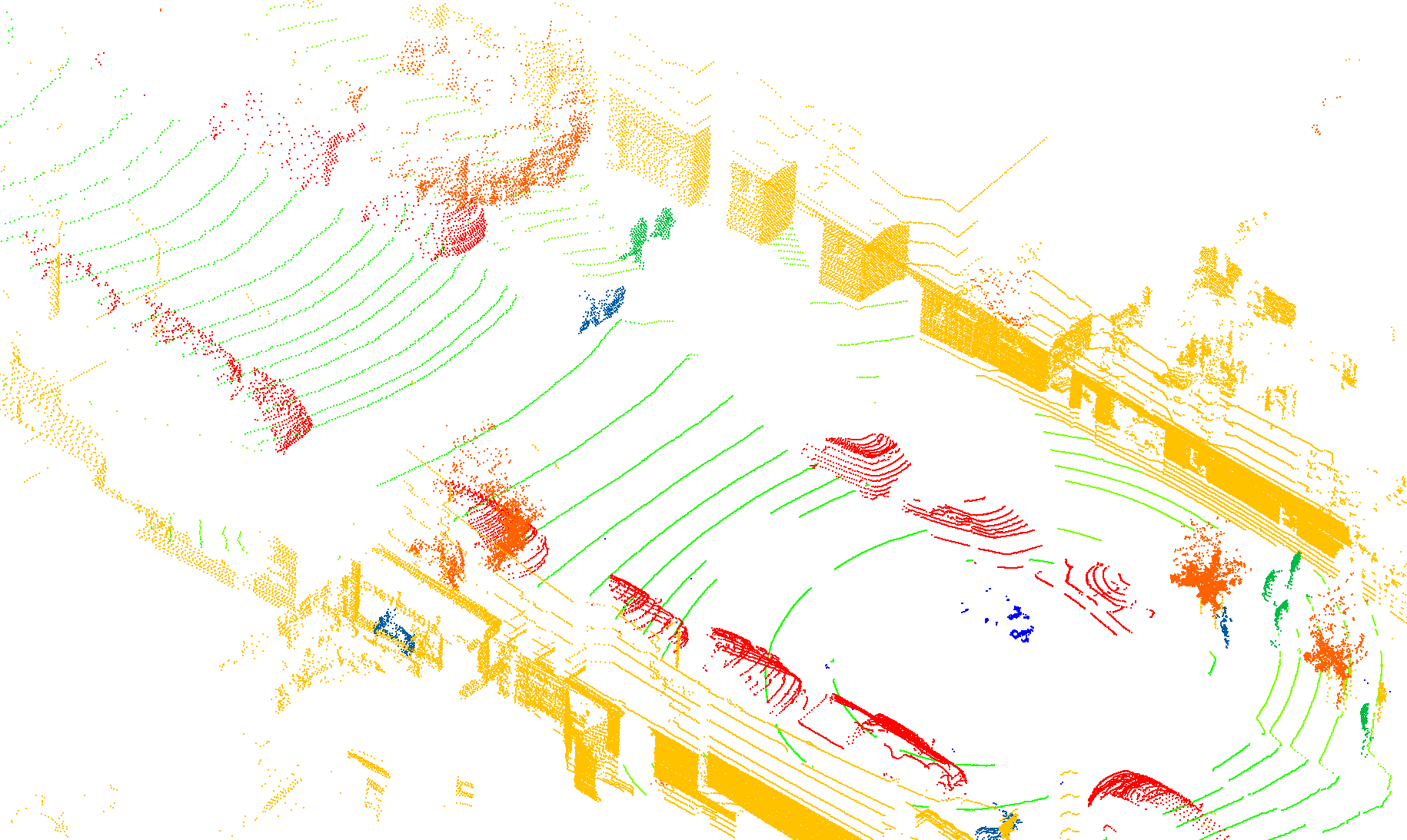}
        }
        \subfloat{
            \includegraphics[width=.25\linewidth]{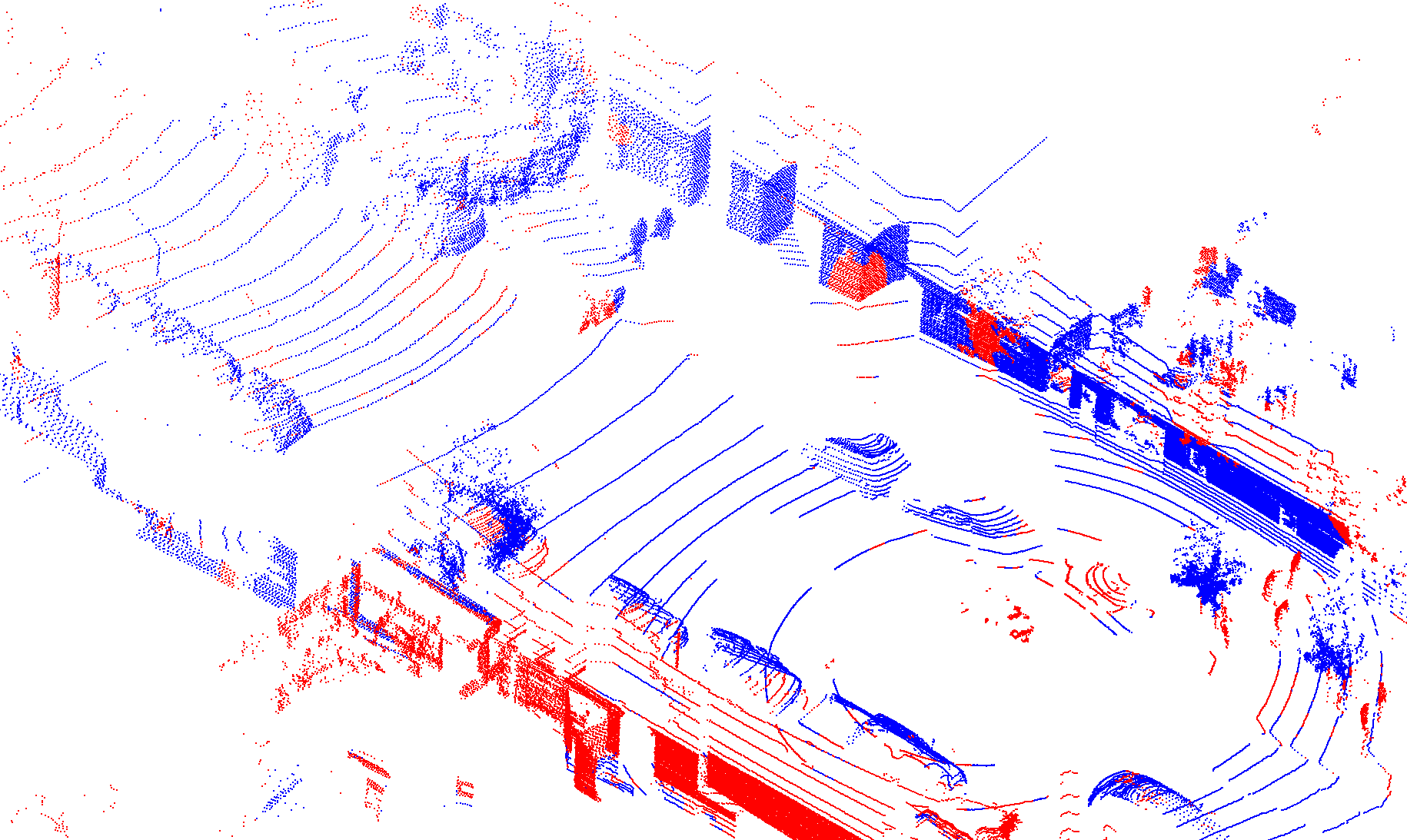}
        } 
        \subfloat{
            \includegraphics[width=.25\linewidth]{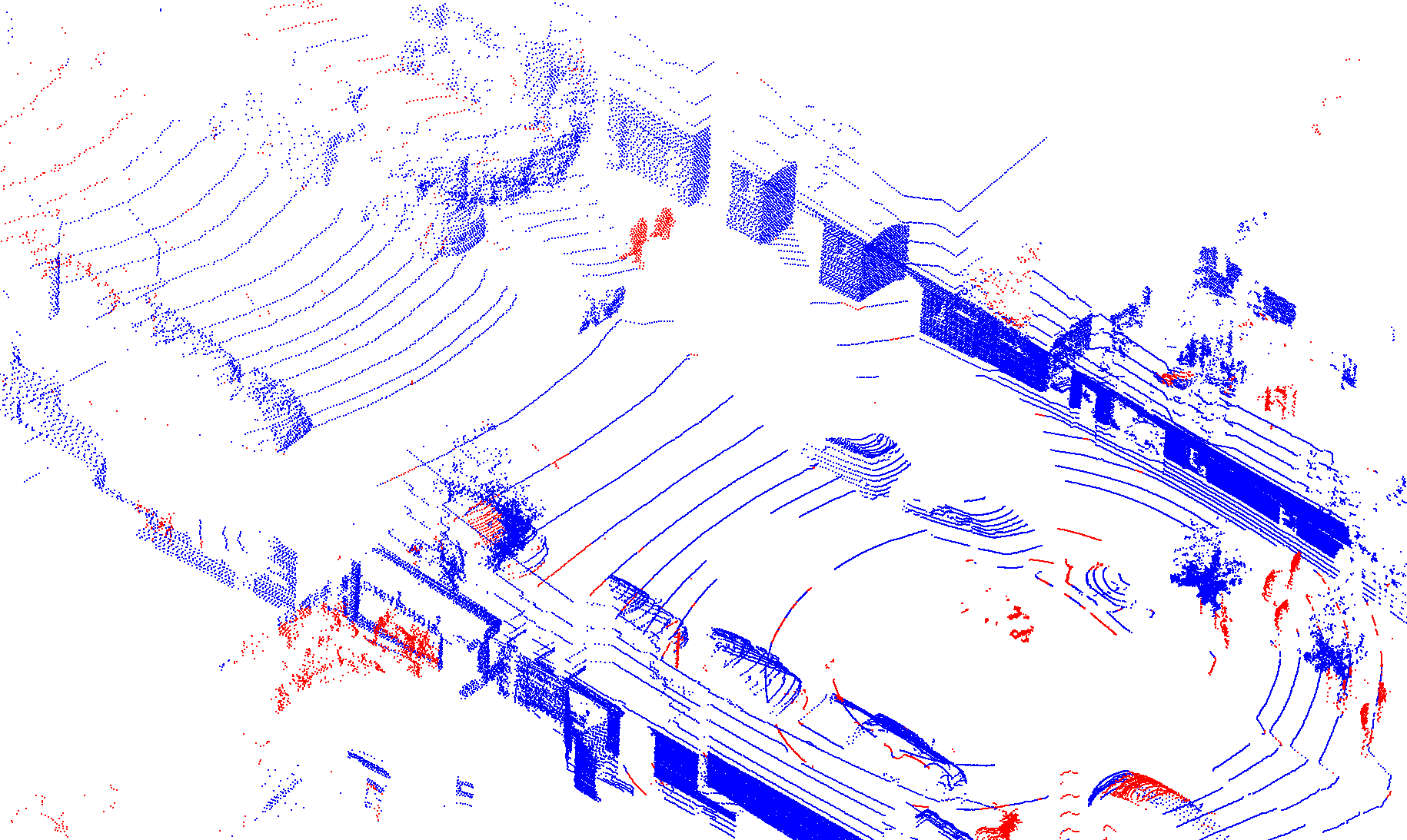}
        }
        \subfloat{
            \includegraphics[width=.25\linewidth]{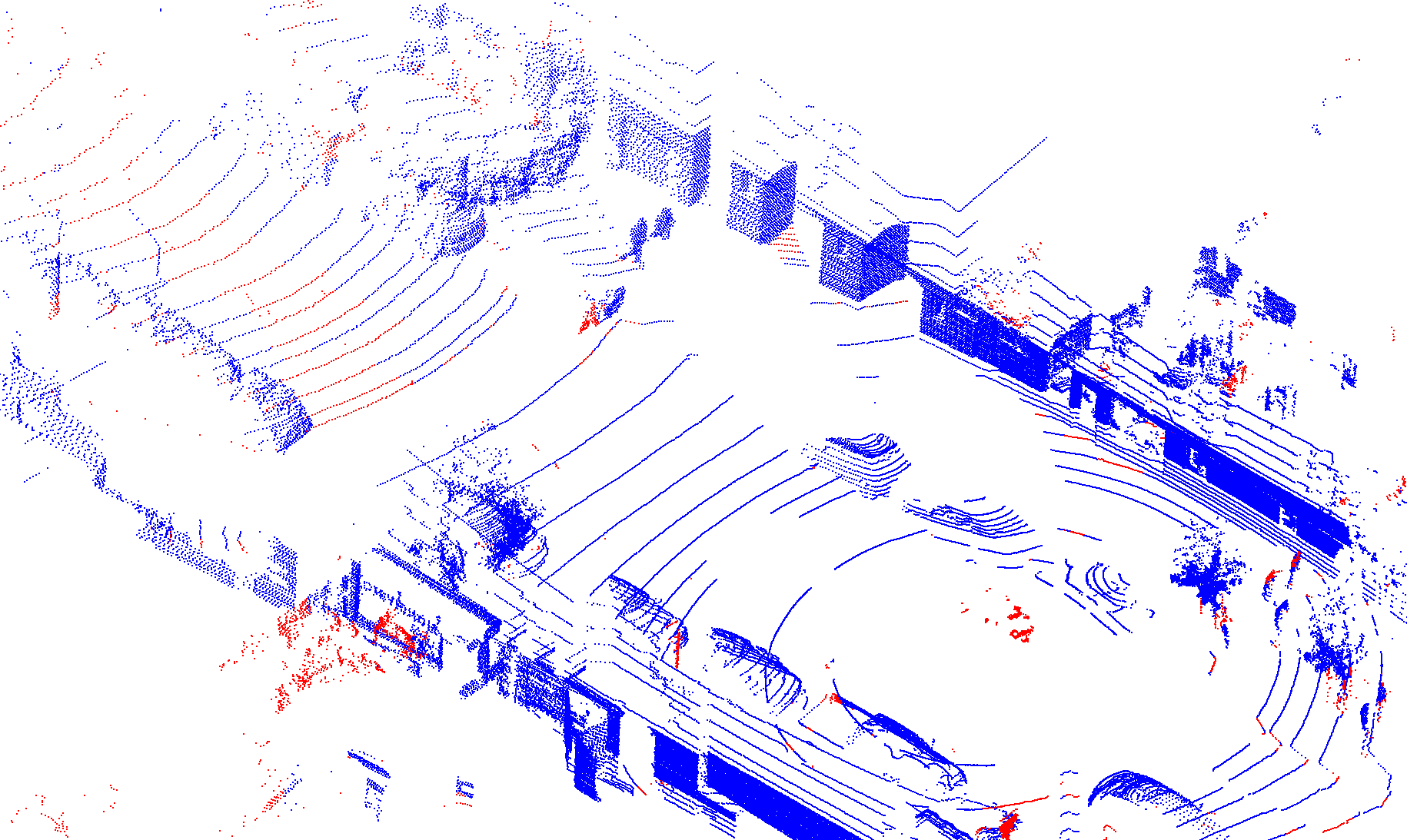}
        } 
        
        \subfloat[Ground Truth]{
            \includegraphics[width=.25\linewidth]{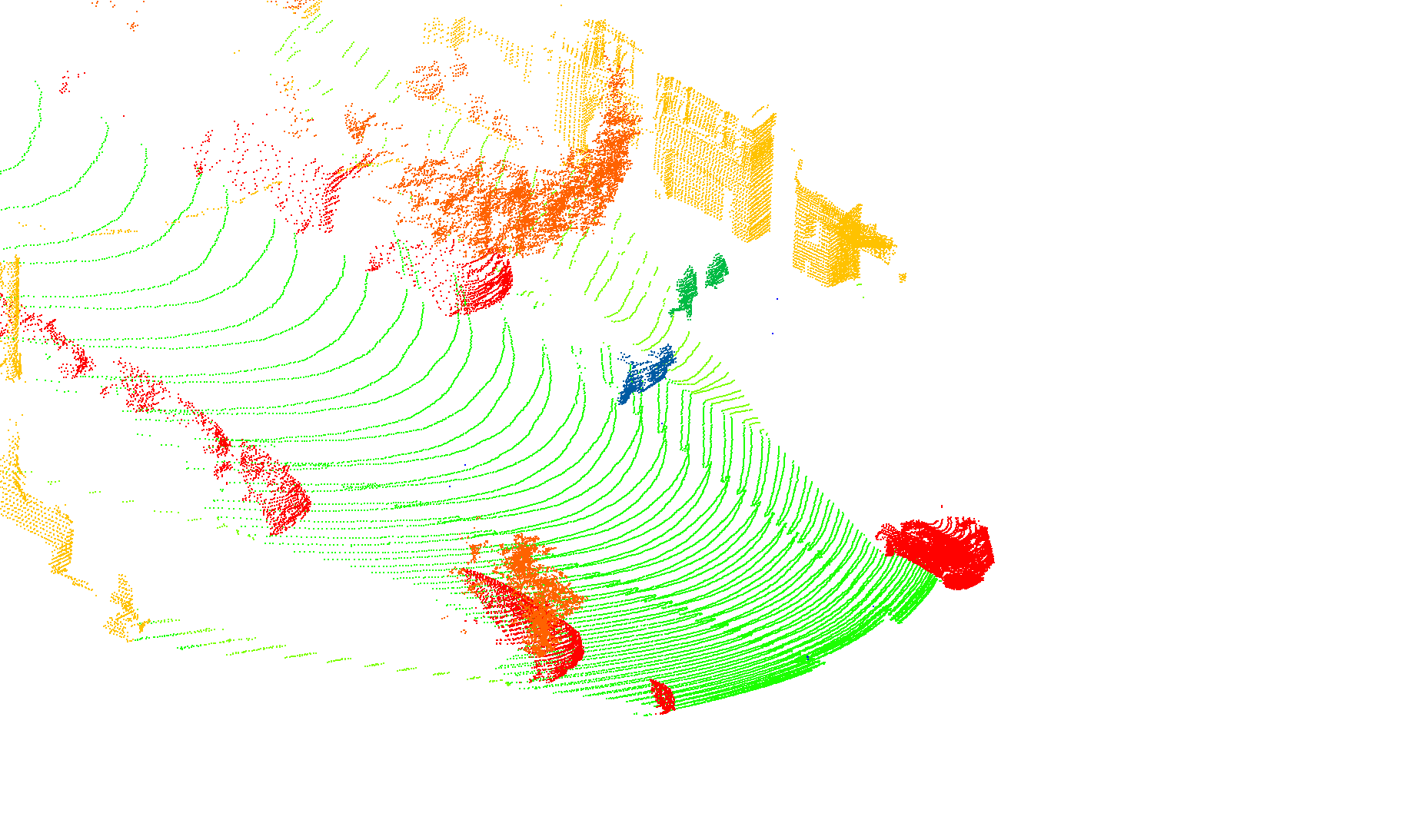}
        }
        \subfloat[KPConv]{
            \includegraphics[width=.25\linewidth]{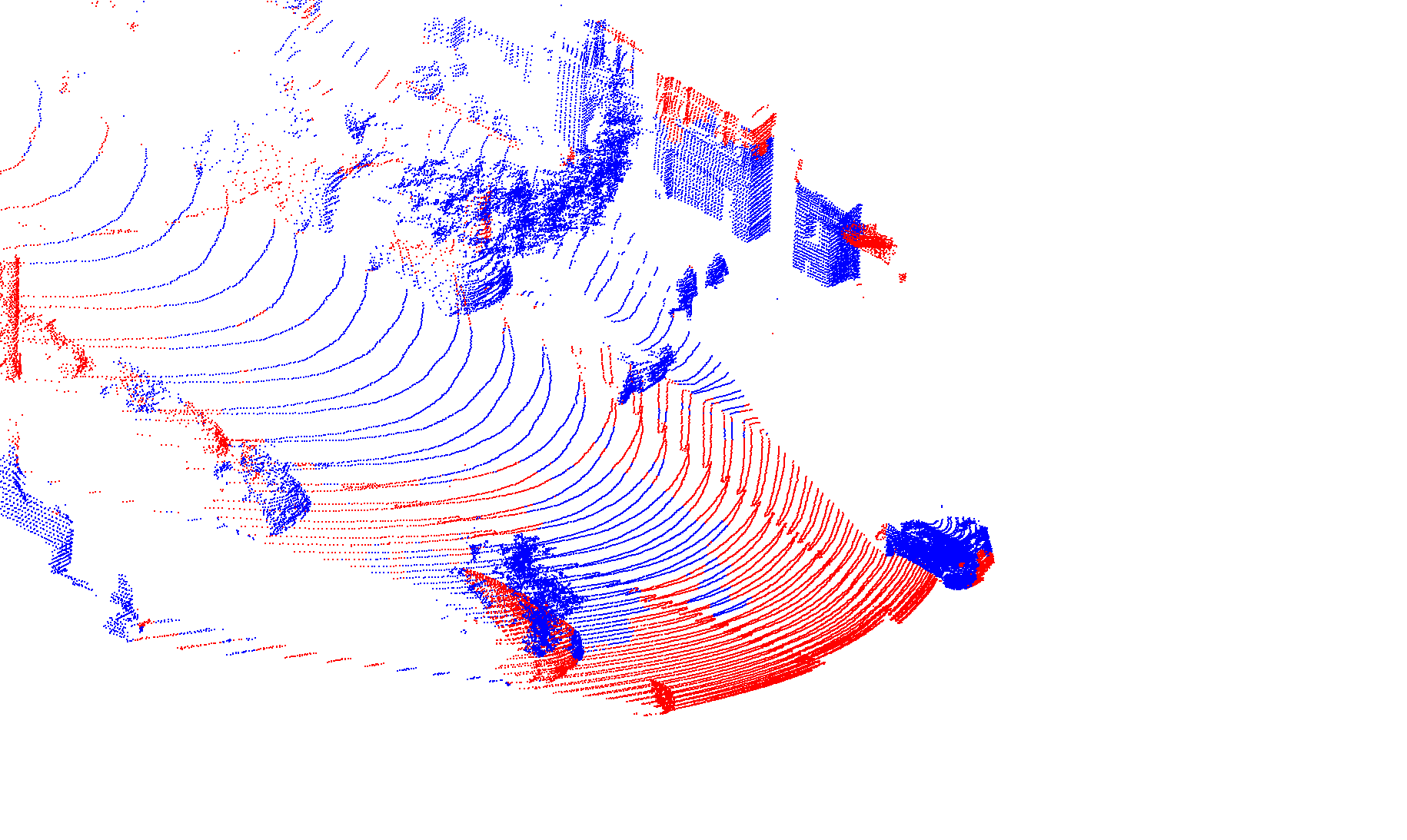}
        } 
        \subfloat[SPVCNN]{
            \includegraphics[width=.25\linewidth]{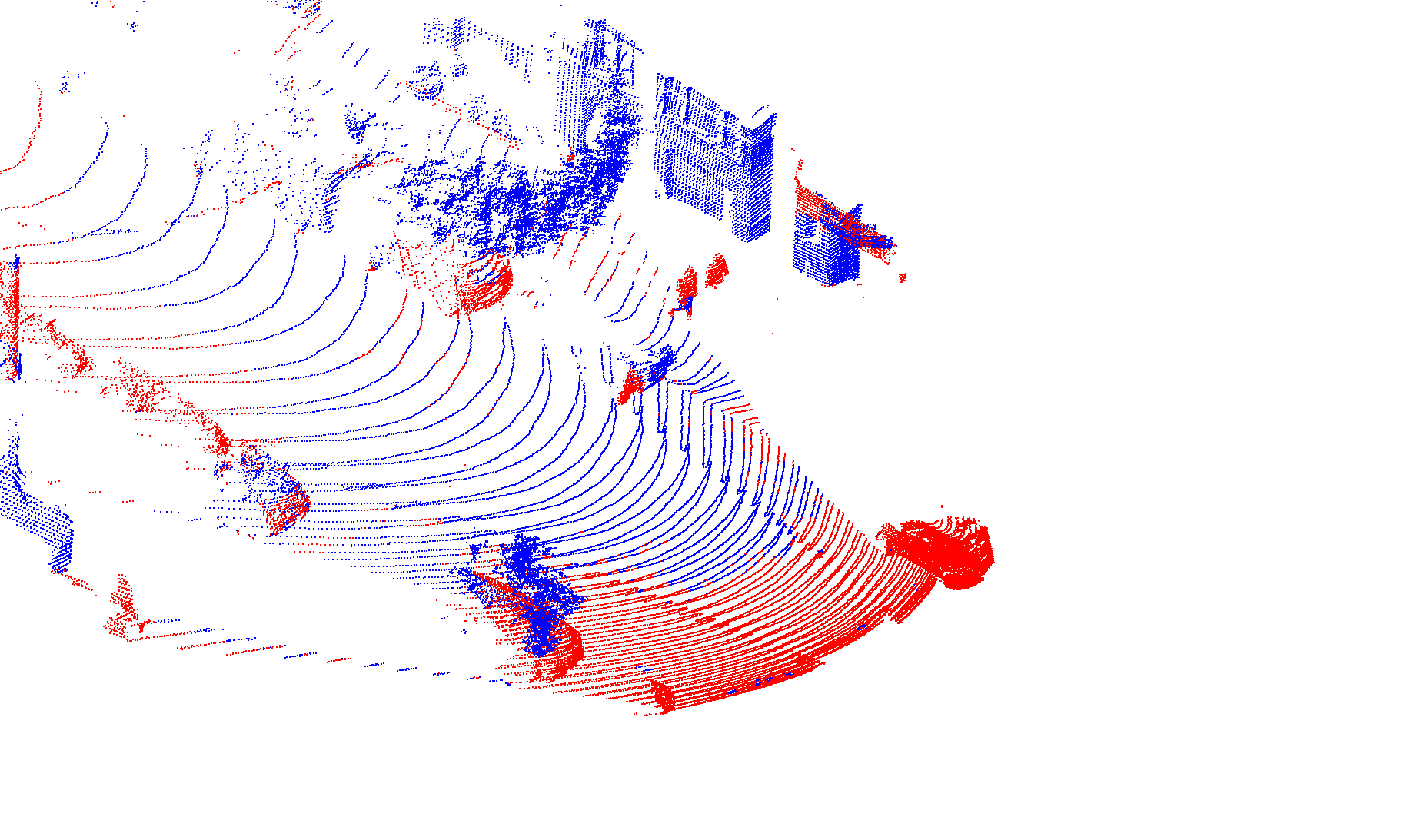}
        }
        \subfloat[3DLabelProp (Ours)]{
            \includegraphics[width=.25\linewidth]{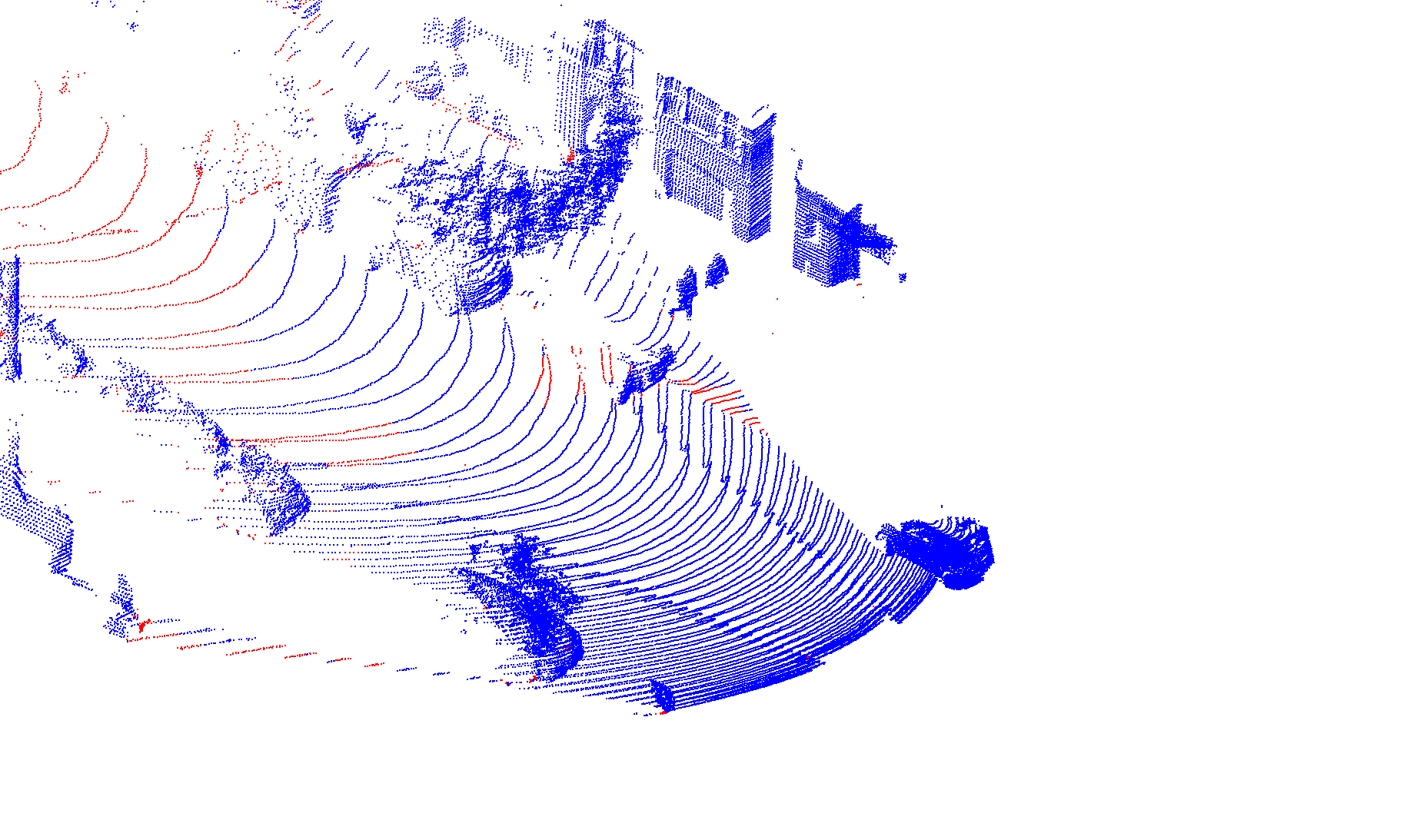}
        } 
        \caption{Qualitative results when trained on SemanticKITTI and tested on Panda64 (top row) and PandaFF (bottom row). For comparison in the same environment, two frames acquired at the same time were chosen for Panda64 and PandaFF. From left to right: Ground truth labels, results from KPConv, results from SPVCNN, results from 3DLabelProp. In blue, points with correct semantic segmentation. In red, errors. \\ For Panda64 we see the ability of 3DLabelProp to properly segment point close to the sensor contrary to SPVCNN where most of its errors lie. The same pattern can be seen for PandaFF where SPVCNN and KPConv have a lot of trouble understanding points with limited neighborhoods, i.e points close to the sensor, whereas 3DLabelProp display the same pattern of errors for Panda64 and PandaFF and has no issue with points with limited neighborhoods.}
\end{figure*}

\begin{figure*}
        \centering

        \subfloat[Ground truth]{
            \includegraphics[width=.25\linewidth]{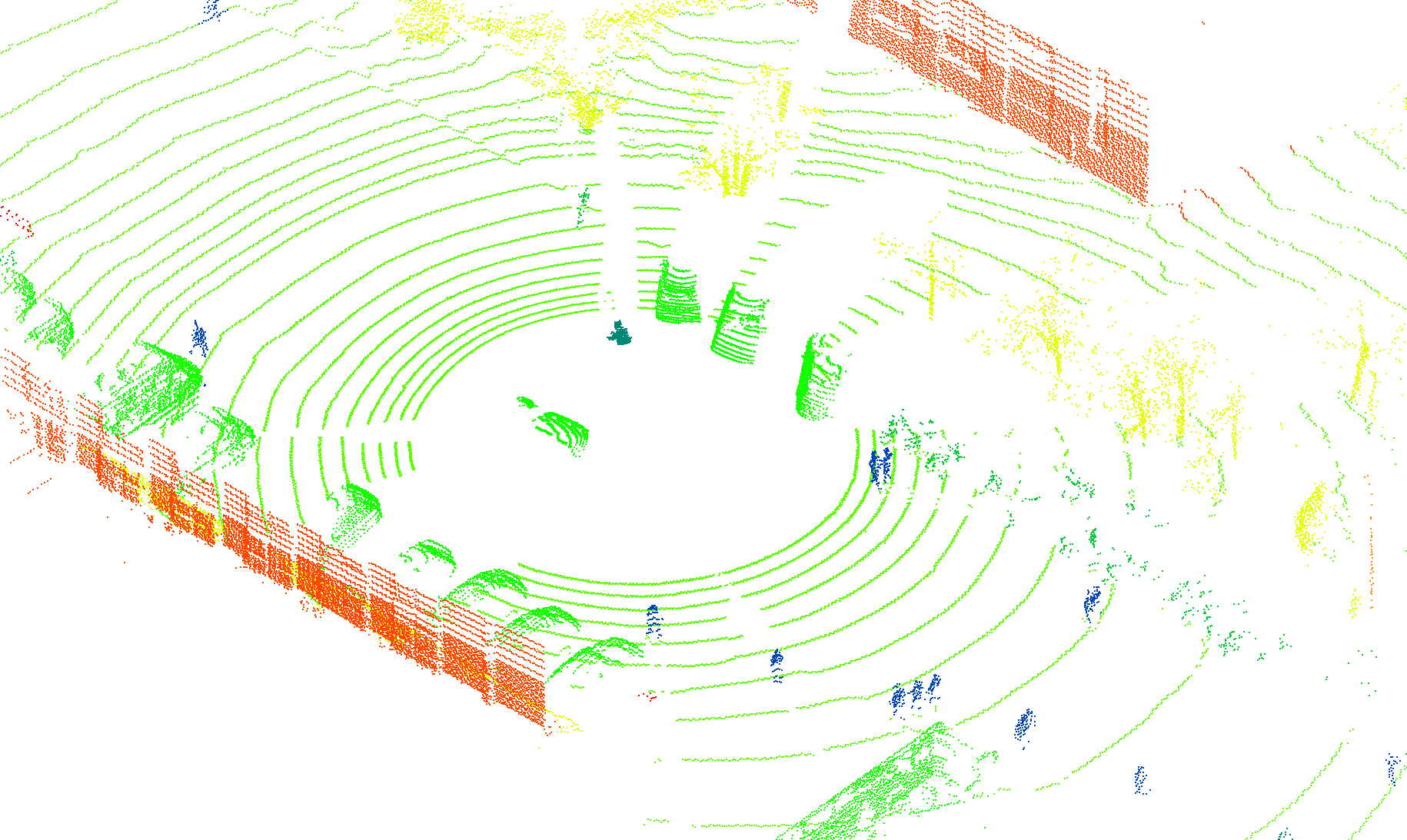}
        }
        \subfloat[KPConv]{
            \includegraphics[width=.25\linewidth]{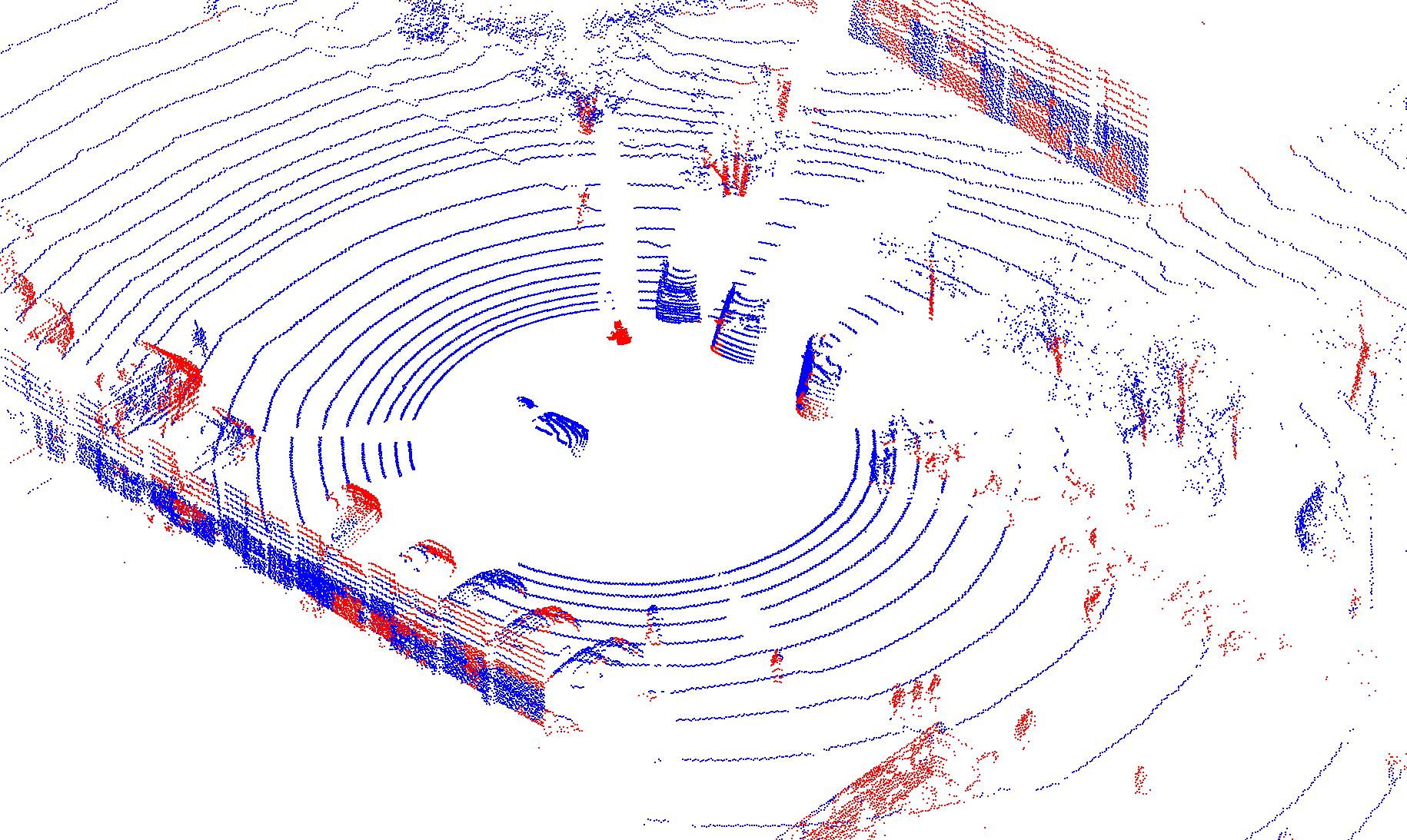}
        } 
        \subfloat[SPVCNN]{
            \includegraphics[width=.25\linewidth]{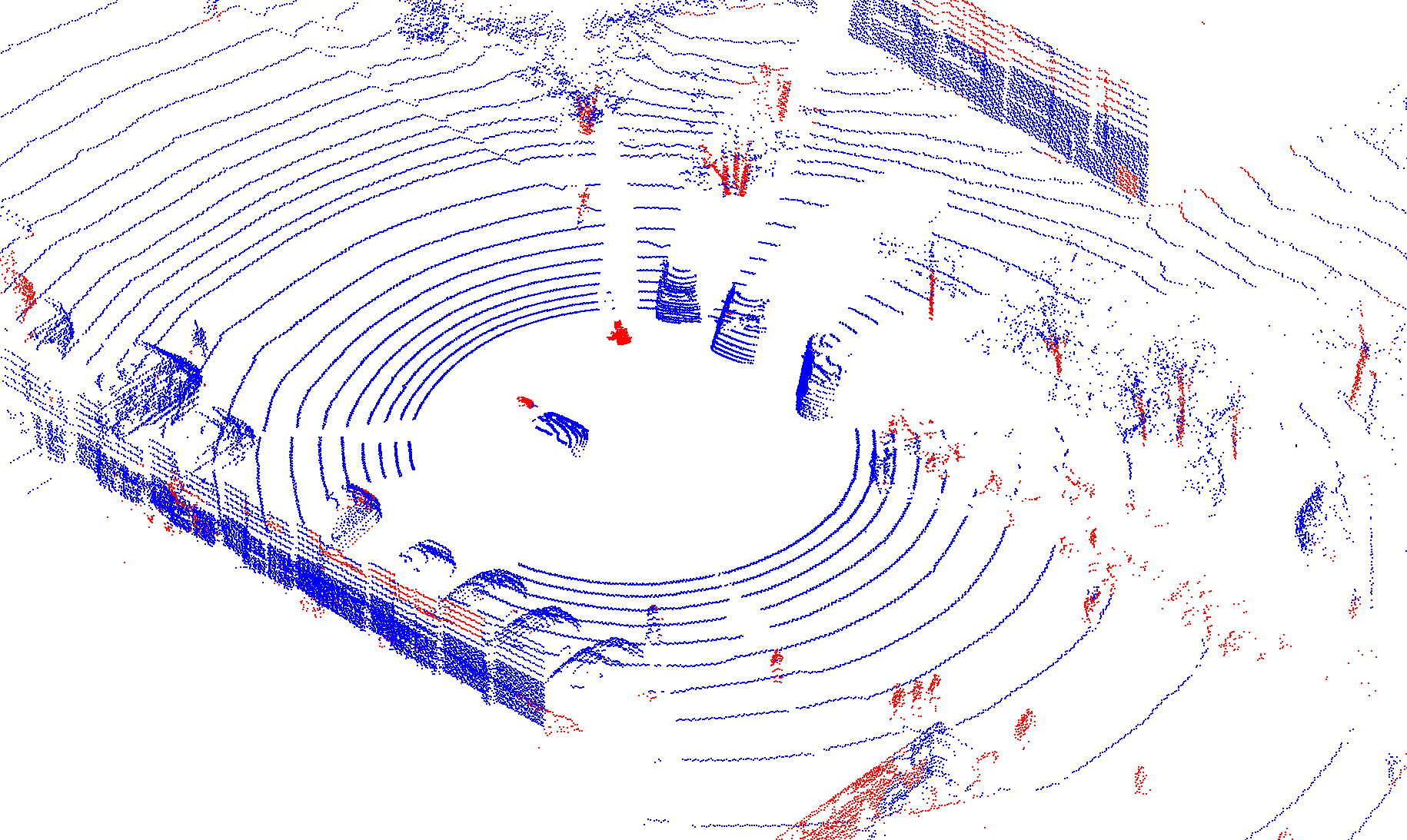}
        }
        \subfloat[3DLabelProp (Ours)]{
            \includegraphics[width=.25\linewidth]{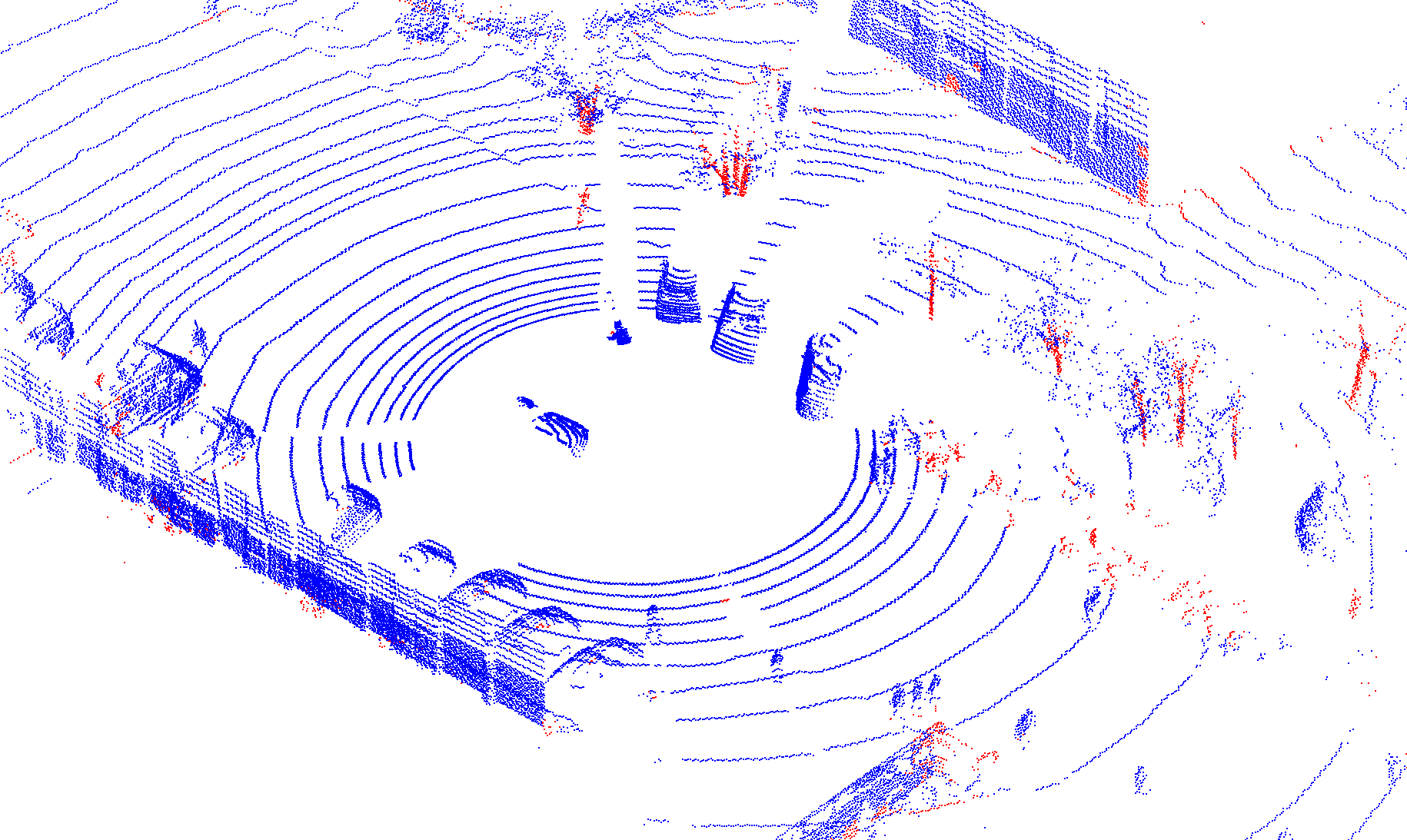}
        } 
        \caption{Qualitative results when trained on SemanticKITTI and tested on SemanticPOSS. From left to right: Ground truth labels, results from KPConv, results from SPVCNN, results from 3DLabelProp. In blue, points with correct semantic segmentation. In red, errors.\\ 
        In these images, we see the ability of 3DLabelProp to properly recognize all the parked cars, in contrary to KPConv. Buildings are also more accurately detected for 3DLabelProp compared to SPVCNN. We see that trunks are systematically badly segmented.}
\end{figure*}

\begin{figure*}
        \centering

        \subfloat[Ground truth]{
            \includegraphics[width=.25\linewidth]{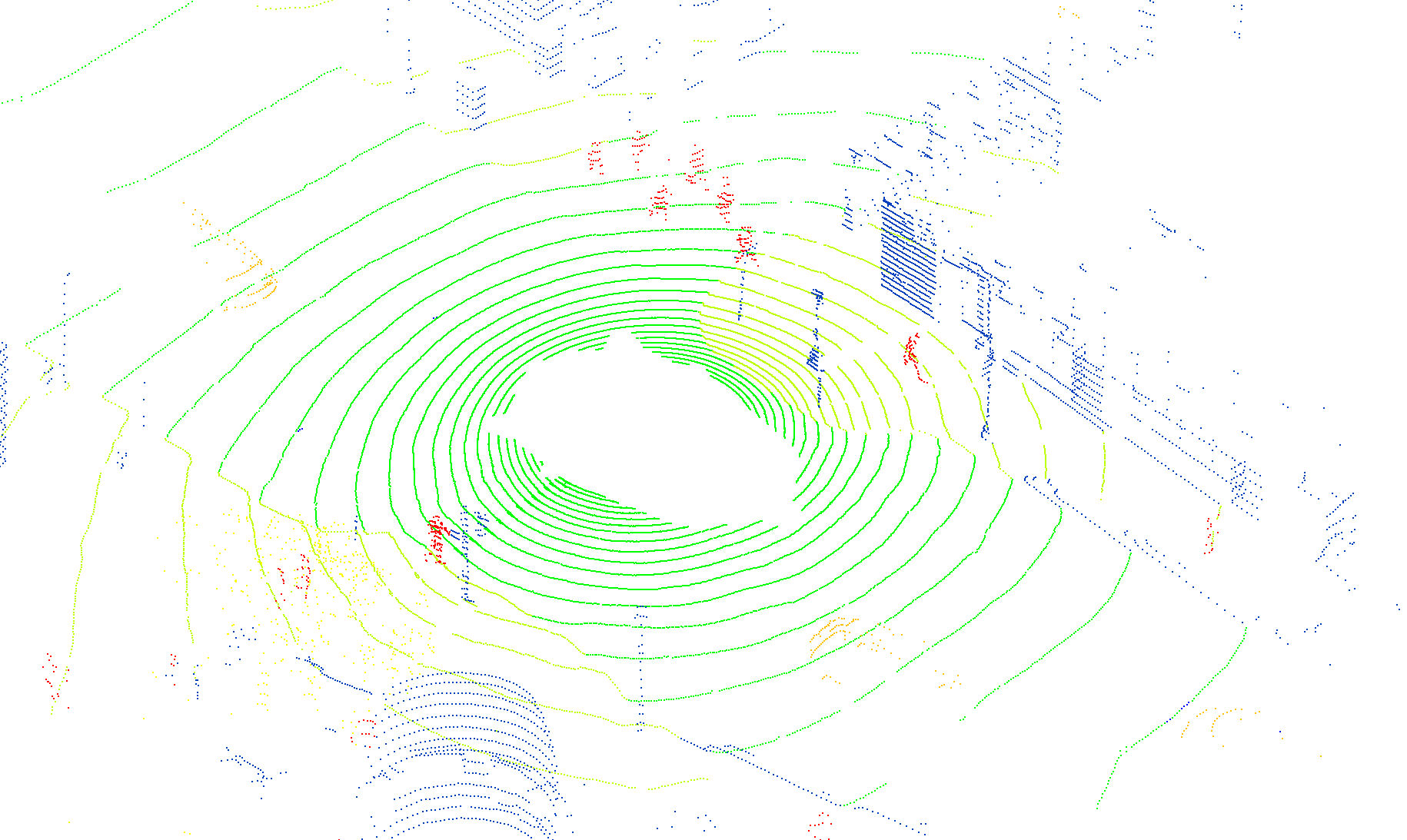}
        }
        \subfloat[KPConv]{
            \includegraphics[width=.25\linewidth]{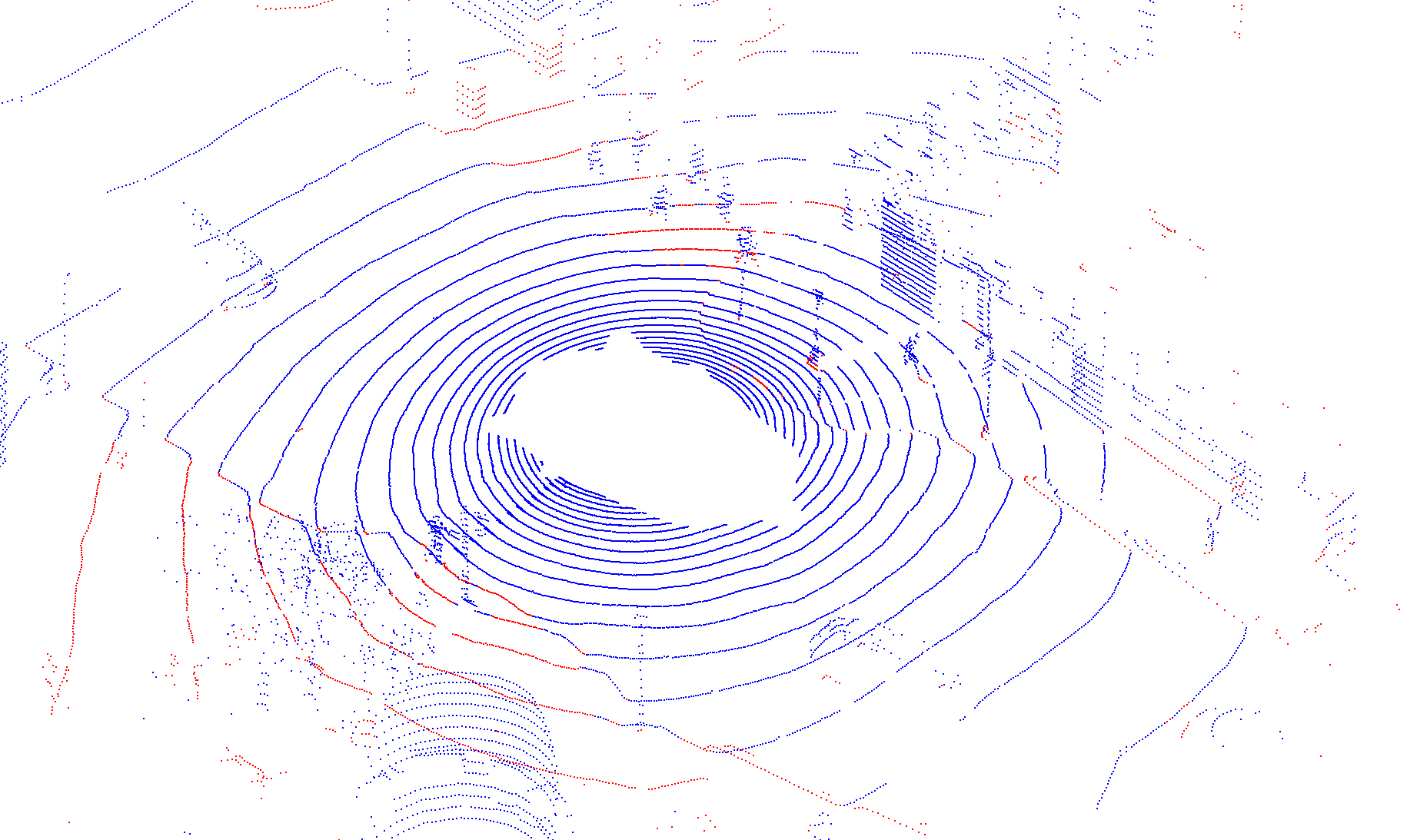}
        } 
        \subfloat[SPVCNN]{
            \includegraphics[width=.25\linewidth]{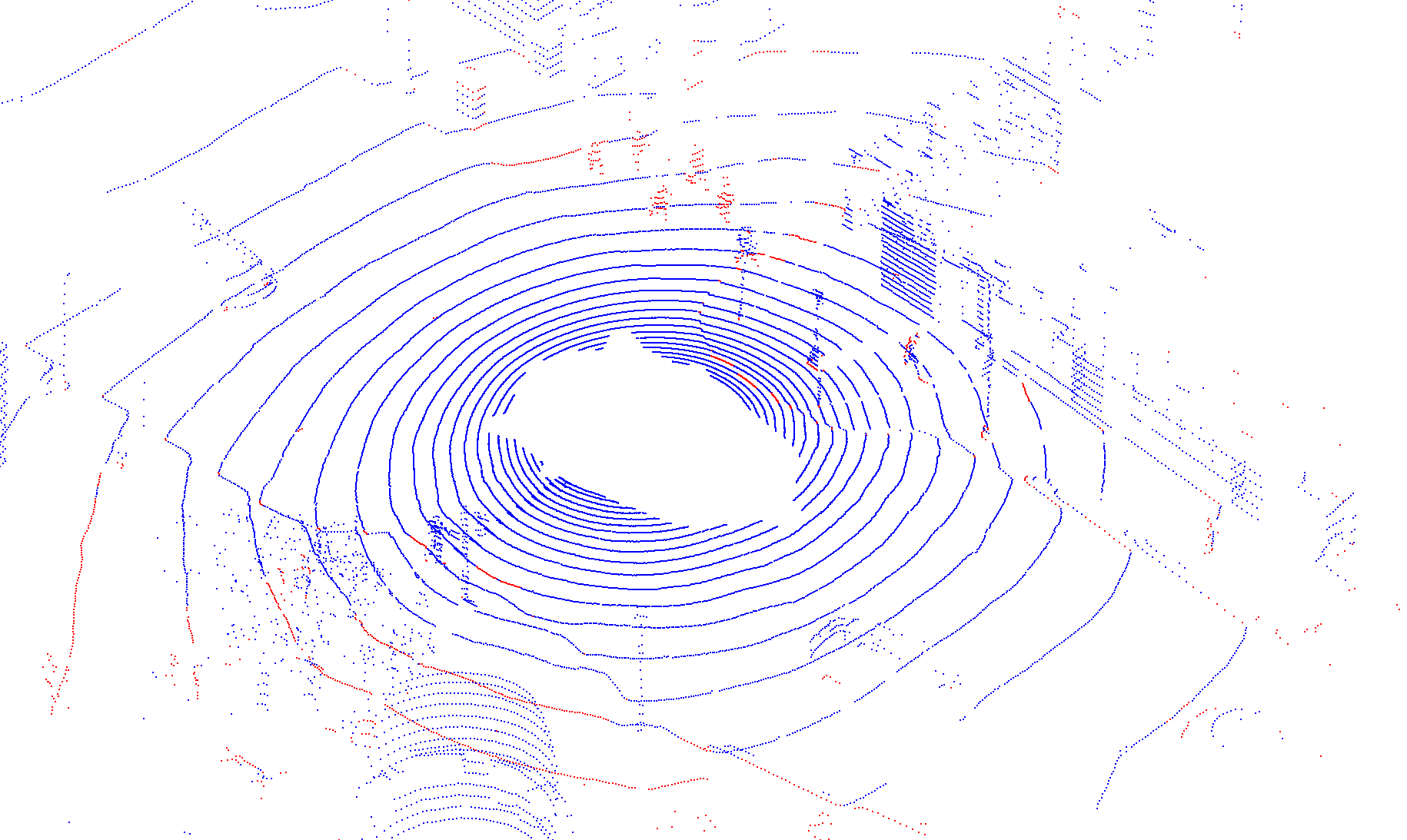}
       }
        \subfloat[3DLabelProp (Ours)]{
            \includegraphics[width=.25\linewidth]{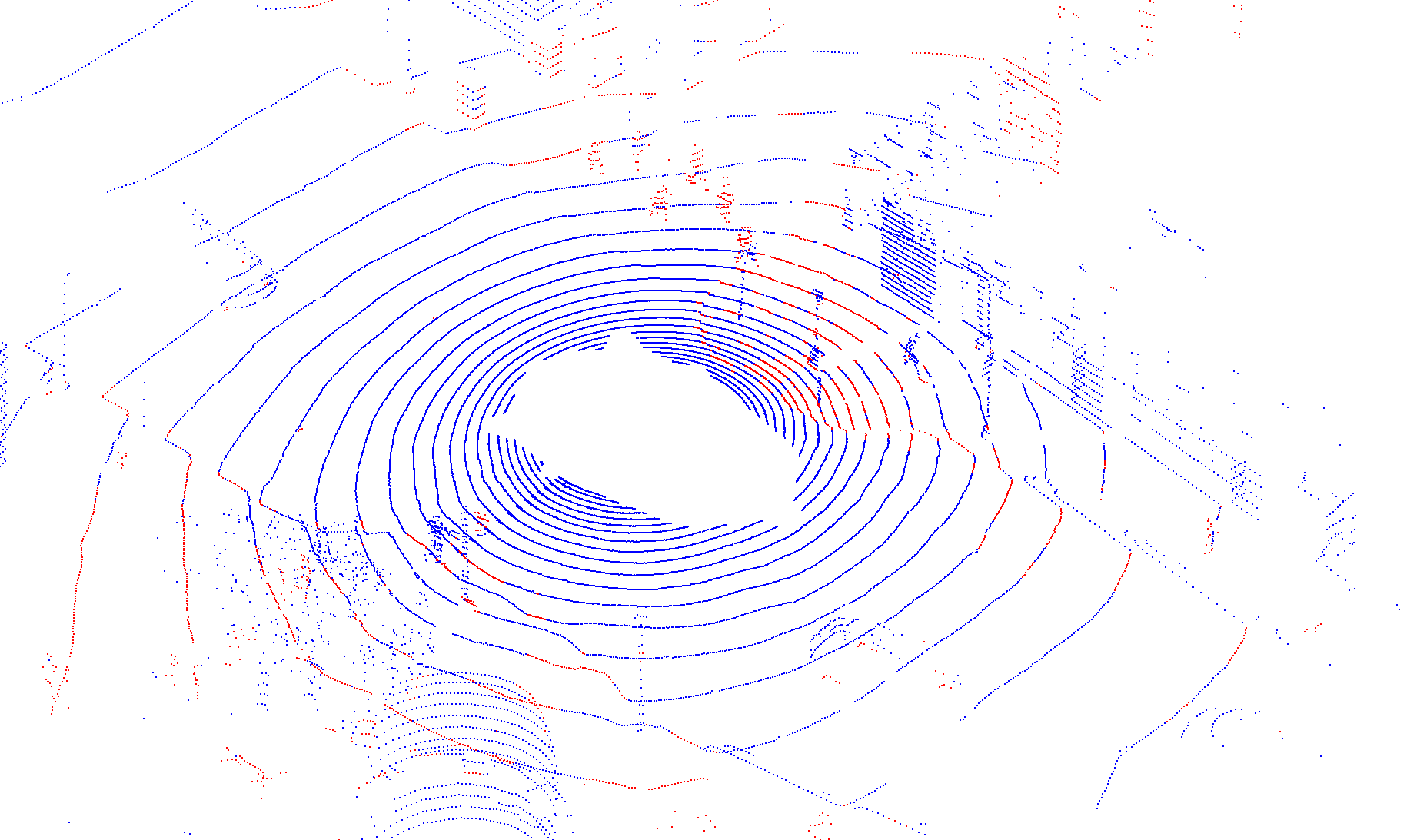}
        } 
        \caption{Qualitative results when trained on SemanticKITTI and tested on nuScenes. From left to right: Ground truth labels, results from KPConv, results from SPVCNN, results from 3DLabelProp. In blue, points with correct semantic segmentation. In red, errors. \\ We see the higher quality of KPConv on pedestrians compared to SPVCNN and 3DLabelProp.}
\end{figure*}

\end{document}